\documentclass{elsarticle}
\usepackage{amsfonts}
\usepackage{dsfont}
\usepackage{amssymb}
\usepackage{amsmath}
\usepackage{bbm}
\usepackage{amsxtra}
\usepackage{graphicx}
\usepackage{psfrag}
\usepackage{mathrsfs}
\usepackage{multirow}
\usepackage{natbib}
\usepackage{subcaption}
\usepackage{tikz}
\usepackage{empheq}
\usepackage[normalem]{ulem}
\usepackage{pdfpages}
\usepackage{color}
\usepackage{longtable}
\usepackage{footmisc}
\usepackage{hyperref}
\usepackage{cleveref}
\usepackage{ulem}
\usepackage[utf8]{inputenc}
\usepackage{lineno}

\newtheorem{algo}{Algorithm}
\newcommand{\bsym}[1]{\boldsymbol{#1}}

\title{Scale bridging materials physics: Active learning workflows and integrable deep neural networks for free energy function representations in alloys}

\author[umme]{G.H. Teichert}
\author[ucsb]{A.R. Natarajan}
\author[ucsb]{A. Van der Ven}
\author[umme,umm,micde]{K. Garikipati\corref{mycorrespondingauthor}}
\ead{krishna@umich.edu}
\cortext[mycorrespondingauthor]{Corresponding Author}

\address[umme]{Department of Mechanical Engineering, University of Michigan}
\address[ucsb]{Materials Department, University of California, Santa Barbara}
\address[umm]{Department of Mathematics, University of Michigan}
\address[micde]{Michigan Institute for Computational Discovery \& Engineering, University of Michigan}

\begin{document}

\begin{abstract}
The free energy plays a fundamental role in theories of phase transformations and microstructure evolution. It encodes the thermodynamic coupling between different fields, such as mechanics and chemistry, within continuum descriptions of non-equilibrium materials phenomena. In mechano-chemically interacting materials systems, even consideration of only compositions, order parameters and strains results in a free energy description that occupies a high-dimensional space. Scale bridging between the electronic structure of a solid and continuum descriptions of its non-equilibrium behavior can be realized with integrable deep neural networks (IDNN) that are trained to free energy derivative data generated by first-principles statistical mechanics simulations and then analytically integrating to recover a free energy density function. Here we combine the IDNN with an active learning workflow to ensure well-distributed sampling of the free energy derivative data in high-dimensional input spaces, thereby enabling true scale bridging between first-principles statistical mechanics and continuum phase field models. As a prototypical material system we focus on Ni-Al. Cahn-Hilliard and Allen-Cahn phase field simulations using the resulting IDNN representation for the free energy density of Ni-Al demonstrate that the appropriate physics of the material have been learned. This work advances the treatment of scale bridging, starting with electronic structure calculations and proceeding through statistical mechanics to continuum physics. Its coupling of Cahn-Hilliard and Allen-Cahn phase field descriptions with nonlinear elasticity through the free energy density ensures a rigorous treatment of phase transformation phenomena.
\end{abstract}


\maketitle

\section{Introduction}
Many continuum models are fundamentally based on an underlying material free energy. For example, the phase field dynamics described by the Cahn-Hilliard and Allen-Cahn equations have at their core, chemical potentials, which are variational derivatives of the total free energy with respect to composition and order parameters, respectively \cite{CahnHilliard1958,Allen1979,Garcia2004,Provatas2011}. Another manifestation is seen in nonlinear elasticity, wherein hyperelastic material models are defined by a strain energy density. The first derivatives of this energy with respect to frame invariant strains define the stresses, and second derivatives give (generally) non-constant elastic moduli. The governing equations for quasi-static elasticity can be derived by extremization of the Gibbs free energy \cite{Marsden1994}. Furthermore, as is obvious for mechano-chemically coupled material systems, cross terms arise among the driving forces, and their correct representation is critical to resolving the dynamics. Due to these fundamental roles, it is important to have a mathematical description of the free energy density that accurately reflects the physics. It is actually important to also control the accuracy of free energy derivatives, since differentiation tends to magnify errors.

Several challenges may arise in constructing such a free energy density function from data. One is rapid fluctuations that may exist in the free energy with respect to its arguments, and that can be difficult to capture. As we have shown, while spline representations prove superior to various polynomial forms \cite{Teichert2017} they too can have limitations \cite{Teichert2019}. Additionally, the data that are calculated or measured are often the derivative of the free energy, rather than the free energy itself. This is typical for statistical mechanics approaches, where the chemical potential is the accessible variable rather than the free energy. In previous work, we introduced a variant on the standard deep neural network (DNN), which we termed an integrable deep neural network (IDNN), to train a chemical free energy density function from chemical potential data, while maintaining the appropriate physics of the system \cite{Teichert2019}. This was done for the free energy density as a function of two variables, namely, composition and an order parameter.

Another potential challenge to training a free energy density function comes from its high-dimensional inputs. DNNs are well suited to handling high dimensional input \cite{Krizhevsky2012,Yong2015,natarajan2018,Teichert2018a,Putin2016}, therefore the greater difficulty lies in the creation of data that are well-sampled in the high-dimensional space. Depending on the method for computing or measuring the free energy or its derivatives, a ``brute force'' approach to sampling the space may be infeasible due to time and cost. Furthermore, the foundations of theoretical descriptions such as statistical mechanics and continuum physics can prove to be at odds in a manner such that the notions of inputs to and outputs of relations can become reversed as the bridge between scales is crossed. An example appears in this work: the computational approach for statistical mechanics takes certain parameter values as input, and returns composition, order parameters, and the chemical potentials as output. However, the continuum thermodynamics view is of free energy densities, and therefore chemical potentials, being outputs and compositions or order parameters as inputs. In an algorithmic setting, therefore, a continuum computation cannot ``demand'' chemical potentials  at chosen composition or order parameter values. This inability to directly choose the values of the inputs as dictated by theory adds another level of complexity to the creation of a well-sampled dataset in higher dimensions.

Active machine learning approaches (active learning) can provide solutions to the needs of data sampling in a high-dimensional space. Active learning algorithms are designed to query for additional data where they would be most useful \cite{Settles2012}. In this work, we employ an error-based active learning routine in concert with an IDNN to sample chemical potential data for a material system with one composition and three order parameters as inputs. Embedded in the active learning routine is an iterative, boot-strapping approach that combines the input-output mapping property of neural networks with a linear (therefore invertible) relation between chemical potentials and auxiliary bias potentials. The resulting workflow also circumvents the difficulty of input-output relations alluded to above. With this constellation of advances, we are able to compute DNN representations of the free energy density function, which is used in phase field simulations to model the growth of precipitates in a Ni-Al alloy.

This work advances the treatment of scale bridging, starting with electronic structure calculations and proceeding through statistical mechanics to continuum physics. The final formulation, coupling the Cahn-Hilliard and Allen-Cahn phase field descriptions with nonlinear elasticity results in a sophisticated treatment of phase transition phenomena and builds on the work of others in the field \cite{thornton12004,thornton22004,Wang2019,LQChen2013,LQChen2014}.

The paper is organized as follows: Section \ref{sec:first_principles} describes the atomistic and statistical mechanics methods used to obtain chemical potential data, using the Ni-Al system as an example. The IDNN is outlined in Section \ref{sec:IDNN}. The active learning workflow, a centerpiece of this communication, is described in Section \ref{sec:active}. The treatment of elasticity by incorporation of strain energy data is described in Section \ref{sec:elas}. The phase field method is outlined in Section \ref{sec:pf}. Workflow and phase field results are presented in Section \ref{sec:results}.
Concluding remarks appear in Section \ref{sec:conclusions}.

\section{Chemical potential data from atomic level models}
\label{sec:first_principles}

\begin{figure}
    \centering
    \includegraphics[width=0.5\textwidth]{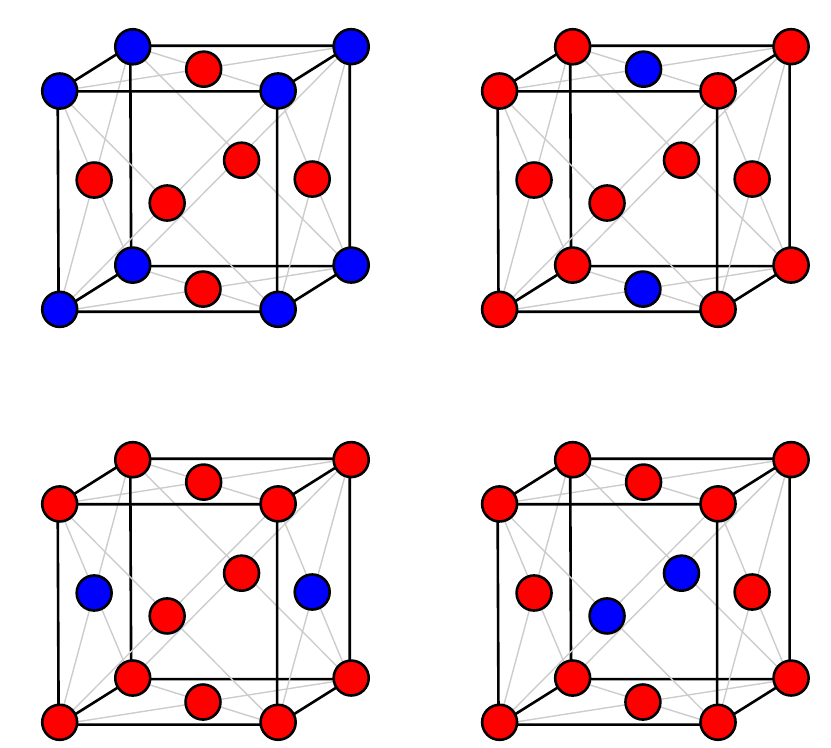}
    \caption{Four variants of the L1$_2$ crystal structure of Ni$_3$Al. The Ni atoms are colored red, and the Al atoms are colored blue.}
    \label{fig:NiAl_var1}
\end{figure}

As a model system we consider Ni-rich Ni-Al alloys, which exhibit interesting order-disorder phenomena on the face-centered cubic (FCC) crystal structure\cite{goiri2016}. At dilute Al concentrations, Ni-Al alloys form an FCC solid solution characterized by disordered arrangements of Ni and Al over the sites of the FCC lattice. At compositions around the Ni$_3$Al stoichiometry, the Ni and Al atoms prefer an ordered arrangement on FCC, adopting the L1$_2$ ordering, which has a lower translational symmetry than the underlying parent FCC lattice. 
While the primitive repeat unit of FCC consists of one site, that of the L1$_2$ ordering has four sites. 
This results in four symmetrically equivalent translational variants of the L1$_2$ ordering as illustrated in Figure \ref{fig:NiAl_var1}. 
The translational variants can coexist and when they impinge on each other, they form an anti-phase boundary. 

The thermodynamic properties of alloys that undergo order-disorder transformations can be calculated with statistical mechanics \cite{Vanderven2018}. This requires a mathematical representation for tracking the instantaneous arrangement of atoms over the sites of the parent crystal, which is realized by assigning an occupation variable $\sigma_i$ to each lattice site $i$ with $\sigma_i = \pm 1$ depending on whether the site is occupied by Ni or Al. The collection of all occupation variables forms the vector $\bsym{\sigma}\in \mathbb{Z}^{n_\text{lat}}$, where $n_\text{lat}$ is the number of lattice sites. The energy of the crystal for any ordering $\bsym{\sigma}$ can be expressed as a polynomial expansion of the occupation variables $\sigma_i$ according to \cite{Sanchez1984,Vanderven2018}
\begin{equation}
    E(\bsym{\sigma}) = E_0 + \sum\limits_{i}E_1^{i}\sigma_i + \sum\limits_{i,j}E_2^{ij}\sigma_i\sigma_j+\sum\limits_{i,j,k}E_3^{ijk}\sigma_i\sigma_j\sigma_k\dots
    \label{eqn:clusterexp}
\end{equation}
where the successive sums on the right hand side extend over all sites, $i$, all distinct pairs of sites $i,j$, all distinct triplets of sites $i,j,k$, etc., leading to the appellation of cluster expansion for this type of representation. The expansion coefficients, $E_0$, $E_1^{i}$, $E_2^{ij}$, etc. can be fit to a training set of energies for different configurations as calculated with a first-principles method such as density functional theory (DFT) \cite{Vanderven2018}. 
The cluster expansion of Equation (\ref{eqn:clusterexp}) can be evaluated rapidly, making it ideally suited for Monte Carlo simulations to calculate thermodynamic averages. 
A cluster expansion Hamiltonian parameterized by Goiri and Van der Ven \cite{goiri2016} was used to describe the effect of configurational ordering in the binary Ni-Al alloy. 

\begin{figure}
    \centering
    \includegraphics[width=0.3\textwidth]{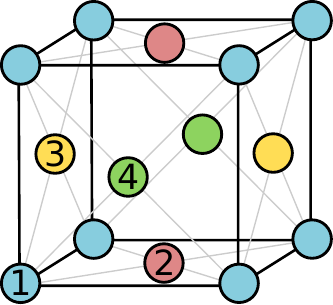}
    \caption{An L1$_2$ structure can be described by the composition of the four sublattice sites numbered here.}
    \label{fig:NiAl_comp}
\end{figure}

\begin{figure}
    \centering
    \includegraphics[width=0.5\textwidth]{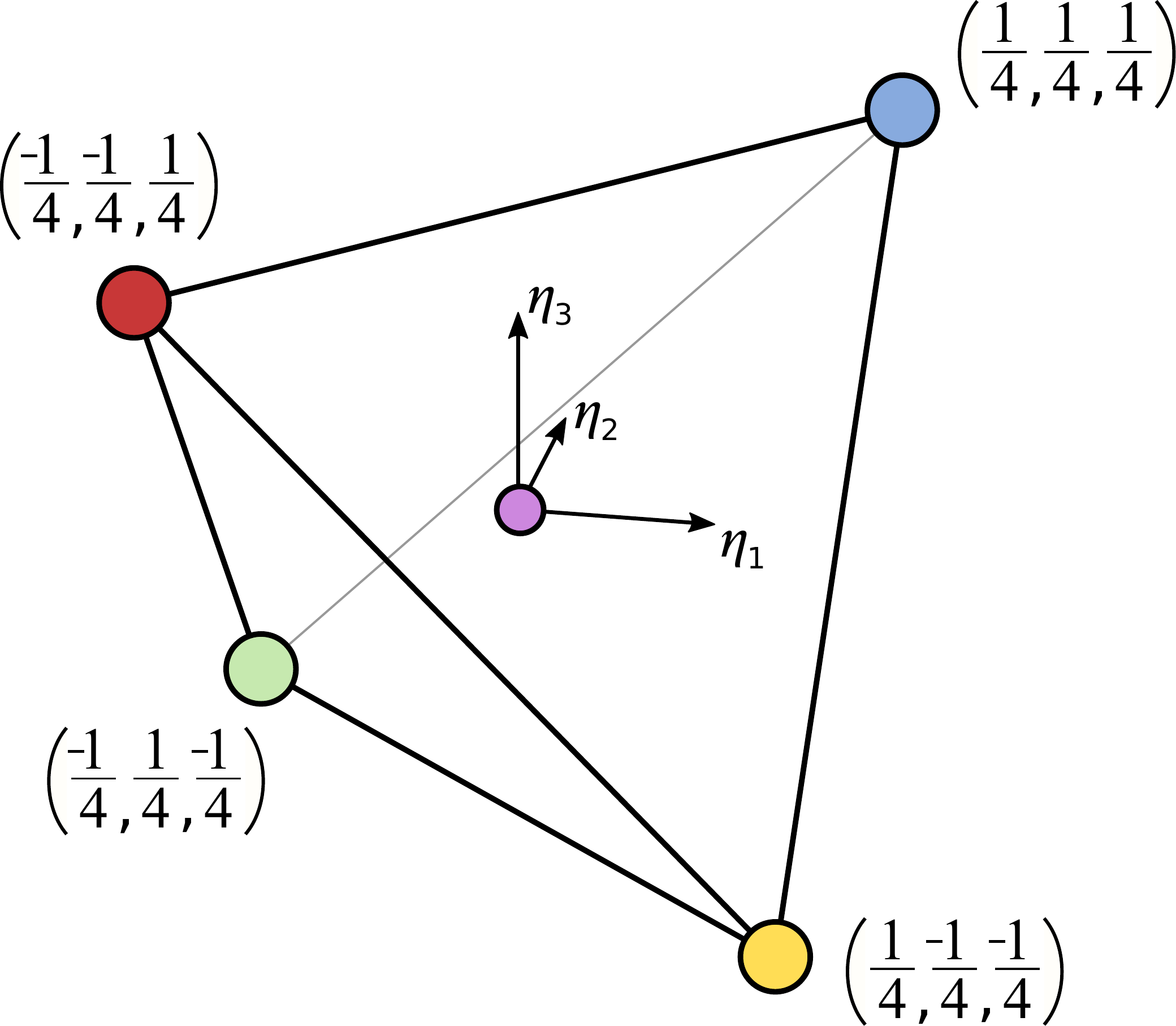}
    \caption{The four perfect L1$_2$ orderings lie on the vertices $(\frac{1}{4},\frac{1}{4},\frac{1}{4})$, $(-\frac{1}{4},-\frac{1}{4},\frac{1}{4})$, $(-\frac{1}{4},\frac{1}{4},-\frac{1}{4})$, $(\frac{1}{4},-\frac{1}{4},-\frac{1}{4})$ of the tetrahedron in the $(\eta_1.\eta_2,\eta_3)$ space. The origin corresponds to a completely disordered state.}
    \label{fig:NiAl_var2}
\end{figure}

The atoms of an alloy in thermal equilibrium constantly fluctuate from one arrangement to another. Nevertheless, the average degree of ordering remains constant at equilibrium. In this context, it is convenient to introduce thermodynamic long-range order parameters \cite{Natarajan2017} that track the equilibrium degree of ordering. 
The degree with which Ni and Al adopt L1$_2$ type ordering can be tracked with average sublattice composition variables $x_i$, $i=1,\ldots,4$, one for each of the four sublattices of the cubic unit cell of L1$_2$ shown in Figure \ref{fig:NiAl_comp}. 
Symmetry arguments then suggest the following linear combinations of the sublattice concentrations for the L1$_2$ ordering
\cite{Braun1997,Natarajan2017}:
\begin{equation}
\begin{split}
    \eta_0 &= \frac{1}{4}\left(x_1 + x_2 + x_3 +x_4\right)\\
    \eta_1 &= \frac{1}{4}\left(x_1 + x_2 - x_3 - x_4\right)\\
    \eta_2 &= \frac{1}{4}\left(x_1 - x_2 - x_3 + x_4\right)\\
    \eta_3 &= \frac{1}{4}\left(x_1 - x_2 + x_3 - x_4\right),
\end{split}
\label{eqn:order_params}
\end{equation}
which can also be expressed using the transformation matrix $\bsym{Q}$:
\begin{equation}
    \bsym{\eta} = \bsym{Q}\bsym{x}
\end{equation}
where 
\begin{equation}
\bsym{Q} = \frac{1}{4}
    \begin{bmatrix}
    1 & 1 & 1 & 1\\
    1 & 1 & -1 & -1\\
    1 & -1 & -1 & 1\\
    1 & -1 & 1 & -1
    \end{bmatrix}
\end{equation}
In this form, the first order parameter, $\eta_0$, tracks the overall composition of the alloy; to emphasize this point, we define the variable $c := \eta_0$ to represent the composition. 
The three remaining order parameters, $\eta_1$, $\eta_2$ and $\eta_3$, measure the degree of long-range order that is commensurate with the periodicity of the L1$_2$ phase. 
They are equal to zero in the completely disordered alloy (since all sublattice concentrations are then equal to each other) and adopt non-zero values when the alloy exhibits average long-range order. 
Furthermore, the three order parameters are able to distinguish between the four translation variants of L1$_2$. 
This is illustrated in Figure \ref{fig:NiAl_var2}, which shows that each translational variant of L1$_2$ (Figure \ref{fig:NiAl_var1}) corresponds to a corner of a tetrahedron in the three dimensional $\eta_1$, $\eta_2$ and $\eta_3$ order-parameter space at a composition $c = \frac{1}{4}$.

Simulating the microstructure evolution of a two-phase mixture of the disordered solid solution and the different translational variants of an ordered phase requires a free energy density, $f$, that is a function of composition (i.e. $c$) and order parameters (i.e. $\eta_1$, $\eta_2$ and $\eta_3$ for the L1$_2$ ordering). 
In the binary Ni-Al alloy, the free energy will have a minimum at the origin of the $\eta_1$, $\eta_2$ and $\eta_3$ space, corresponding to the disorder phase, at compositions where the solid solution is stable. The energy landscape will also have four minima related by symmetry in the vicinity of the translational variants of L1$_2$ in $\eta_1$, $\eta_2$ and $\eta_3$ space at compositions close to the Ni$_3$Al stoichiometry.
Since the free energy density, $f(c,\eta_1,\eta_2,\eta_3)$, is a continuous curve, there will be regions in $c,\eta_{1},\eta_{2},\eta_3$ space where $f$ has negative curvatures. In these regions the alloy is unstable with respect to ordering and/or composition fluctuations.  

Each order parameter, $\eta_i$, has a conjugate ``chemical potential'', $\mu_i$, that can be derived from the free energy density, $f$. If, as in this work, $f$ is the homogeneous part of the free energy density, the corresponding chemical potentials are $\mu_0 = \partial f/\partial c$ and $\mu_i = \partial f/\partial \eta_i$. In Monte Carlo simulations of materials systems, which we adopt to extract the chemical potentials, it is easier to control $\mu_i$ than the order parameters $\eta_i$, since the latter are related to the thermodynamic averages of sublattice concentrations. A difficulty, however, emerges in regions where the free energy density has negative curvatures. To access these regions, biased Monte-Carlo simulations \cite{Natarajan2017} with additional bias parameters $\phi_i$ and $\kappa_i$, $i=0,\ldots,3$ are used. The bias parameters are then the inputs to the Monte Carlo simulations, which allow computation of the statistical averages of the order parameters $\langle\eta_i\rangle$, $i=0,\ldots,3$. The bias parameters and statistical averages are related to the derivative of the free energy per atom, $f(c,\eta_1,\eta_2,\eta_3)$ through the following:

\begin{equation}
\begin{split}
        \mu_0 &:= \frac{\partial f}{\partial c}\Big|_{\langle\bsym{\eta}\rangle} = -2\phi_0(\langle c\rangle-\kappa_0) \\
        \mu_i &:= \frac{\partial f}{\partial \eta_i}\Big|_{\langle\bsym{\eta}\rangle} = -2\phi_i(\langle \eta_i\rangle-\kappa_i), \qquad i=1,\ldots,3 
\end{split}
\label{eqn:mu_bias_params}
\end{equation}

The cluster expansions of Equation (\ref{eqn:clusterexp}) and Monte Carlo statistical mechanics calculations were performed with the \texttt{CASM} code \cite{Casm,Vanderven2018,thomas2013,Puchala2013}. The resulting statistical averages were used to calculate the free energy derivative data.

\section{Integrable deep neural networks}
\label{sec:IDNN}
As explained above, the atomic models directly provide data as derivatives of the free energy density. However, for reasons driven by physics-constrained modelling that were explained in the Introduction, we seek to represent the free energy itself in addition to its derivatives. For such purposes, we have previously introduced the notion of an integrable deep neural network (IDNN) \cite{Teichert2019}. IDNNs are trained to derivative data and can be analytically integrated to recover the antiderivative function (e.g. the free energy density). We summarize their mathematical basis and construction here, and refer the reader to the original work \cite{Teichert2019} for details.

\begin{figure}[tb]
    \centering
    \includegraphics[width=0.75\textwidth]{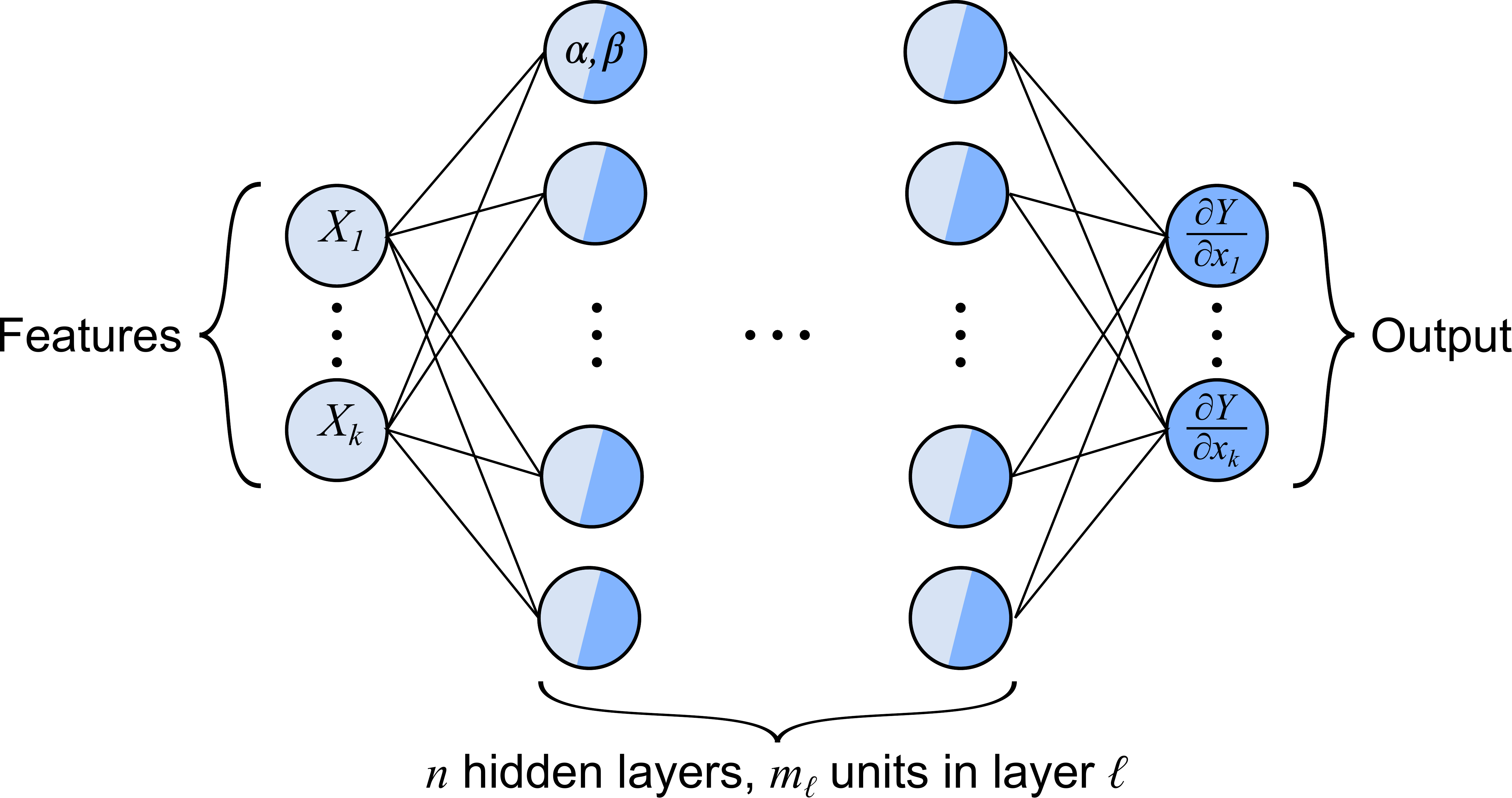}
    \caption{Schematic of an integrable deep neural network (DNN).}
  \label{fig:IDNN}
\end{figure}

Mathematically, the IDNN is constructed by differentiating a standard deep neural network (DNN) by each of its inputs, $X_k$ (see a schematic in Figure \ref{fig:IDNN}). The following equations describe the structure of a standard DNN with $n$ hidden layers, where $\bsym{W}_\ell$, $\bsym{b}_\ell$ are the weight matrix and bias vector of hidden layer $\ell$, $g$ is the activation function, $a_\ell$ and $z_\ell$ are intermediate vector values at each layer, and $Y$ is the DNN output:
\begin{equation}
\begin{split}
    \bsym{z}_\ell &= \bsym{b}_\ell + \bsym{W}_\ell\bsym{a}_{\ell-1}\\
    \bsym{a}_\ell &= g(\bsym{z}_\ell)\\
    Y &= \bsym{b}_{n+1} + \bsym{W}_{n+1}\bsym{a}_{n}
\end{split}
\end{equation}
After differentiation, additional equations arise to describe the IDNN, which is represented by $\partial Y/\partial X_k$:
\begin{equation}
\begin{split}
    \frac{\partial \bsym{a}_\ell}{\partial X_k} &= g'(\bsym{z}_\ell)\odot\left(\bsym{W}_\ell\frac{\partial \bsym{a}_{\ell-1}}{\partial X_k}\right)\\
    \frac{\partial Y}{\partial X_k} &= \bsym{W}_{n+1}\frac{\partial \bsym{a}_n}{\partial X_k}
\end{split}
\end{equation}
where the operator $\odot$ denotes element-wise multiplication. Note that both the activation function and its derivative are used in the IDNN. If the activation function is chosen to be the softplus function, $g(X) := \ln(1 + e^X)$, its derivative, $g'(X) = 1/(1 + e^{-X})$, is also a common activation function, namely the sigmoid--also called the logistic--function. Note that though the IDNN, $\partial Y/\partial X_k$, and its associated DNN, $Y$, have different structures, they share the same weights and biases. It is this fact that creates the derivative/integral relationship between the IDNN and DNN. Of relevance to implementation, the integration to obtain $Y$ is available for no extra training.

Using modern deep learning libraries, an IDNN can simply be defined by constructing a standard DNN, then applying a gradient operator to the output. For a given set of inputs and derivative data $\{(\bsym{\hat{X}}_\theta,\bsym{\hat{y}}_\theta)\}$, the mean square error of the DNN gradient (i.e. the IDNN) and the chemical potential data is minimized over the space of weights and biases, as represented by the following:
\begin{align}
\bsym{\hat{W}},\bsym{\hat{b}} = \underset{\bsym{W},\bsym{b}}{\mathrm{arg\,min}}\,\sum_{k=1}^n\mathrm{MSE}\left(\frac{\partial\bsym{Y}(\bsym{X},\bsym{W},\bsym{b})}{\partial X_k}\Big |_{\bsym{\hat{X}}_\theta},\hat{y}_{\theta_k}\right)
\label{eq:IDNN-Wb}
\end{align}
The resulting trained standard DNN gives the integrated DNN.

\section{Active learning workflow}
\label{sec:active}
It is desirable to have a free energy derivative that is uniformly sampled in the space of order parameters for use in mesoscale models. However, Monte-Carlo techniques use the bias parameters $\phi_{i}$ and $\kappa_{i}$ as input, with the order parameter values emerging as thermodynamic averages from the simulations. The bias parameters are related to the chemical potentials and order parameters through Equation (\ref{eqn:mu_bias_params}). Typically, in biased Monte-Carlo simulations the bias curvature, $\phi_{i}$ is held constant, while $\kappa_{i}$ values are varied. 

Na\"{i}ve sampling of the $\kappa_i$ parameters can lead to under- and over-sampling of some regions in the order parameter space. The uniformity of sampling can be improved by creating and using a surrogate model, $\widehat{\bsym{\mu}}(\bsym{\eta})$, to predict which values of $\kappa_i$ will give uniform sampling in the $\bsym{\eta}$ space. In our treatment, the surrogate $\widehat{\bsym{\mu}}(\bsym{\eta})$ is an IDNN.

While all values of $\kappa_i$ are physically valid, some values are more relevant than others. It is not initially apparent what the relevant range of $\kappa_i$ values should be. However, physically valid values for each sublattice composition $x_i$ lie in the range $[0,1]$. Therefore, instead of using the $\kappa_i$ to define the domain of the search space, we sample from the sublattice composition space. We impose the additional constraint that $c \leq 0.25$, since the Ni-Al system transitions from FCC to BCC for $c > 0.25$. As discussed in the Introduction, the endpoint goal of the scale-bridging framework for materials physics is continuum models governed by partial differential equations (PDEs), including those of phase field models and nonlinear elasticity. Given that the PDEs defining the phase field model are written in terms of the composition and order parameters, we pose the problem in terms of $\bsym{\eta}$.

For each iteration of the workflow, we perform a global sampling from the sublattice composition space using Sobol$^\prime$ sequences--a choice made because of their space-filling and noncollapsing properties \cite{Sobol1967,Bratley1988,Bessa2017}. The sublattice composition values are converted to order parameter values with Equation (\ref{eqn:order_params}). These are used as input to the surrogate model, which gives a prediction for the chemical potentials and, using Equation (\ref{eqn:mu_bias_params}), the associated $\kappa_i$ bias parameters. With these $\kappa_i$ values as input, the cluster expansions and Monte Carlo computations (within the \texttt{CASM} platform) return a set of composition and order parameter values, $\eta_i$, with their corresponding chemical potentials, $\mu_i$, for $i = 0,\dots 3$.

Once the dataset is updated, the IDNN is trained using all of the chemical potential data. After training is complete for the current iteration, the active learning component of the workflow takes place. The pointwise training error is evaluated for the IDNN using only the data points from the most recent global sampling. The data points are sorted according to error. The $N$ data points giving the highest error are used to identify regions of space that need more data. Additionally, the appearance of energy wells in the surface are of interest, since they correspond with the material phases. These energy wells are identified by evaluating the Hessian of the free energy surface for sampled points and selecting points with a positive definite Hessian and a low gradient norm (within some tolerance of zero). Random points near these data with either high error or within an energy well are used to define a local sampling of order parameter values. As before, the IDNN as the surrogate model and associated equations provide $\kappa_i$ values that become input to Monte Carlo calculations via \texttt{CASM}, resulting in an updated dataset and concluding the iteration.

\begin{figure}
    \centering
    \includegraphics[width=0.8\textwidth]{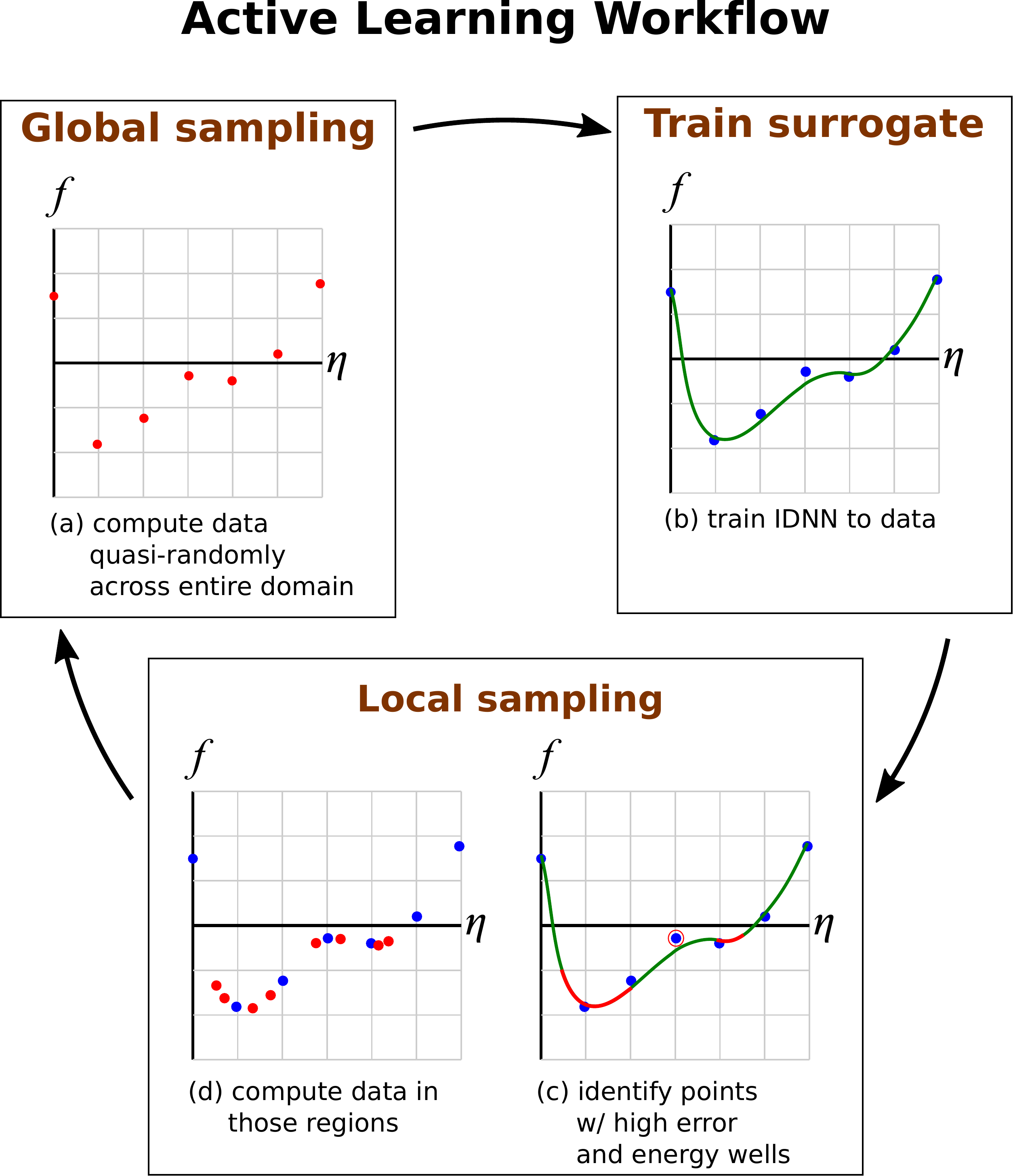}
    \caption{Schematic of the active learning workflow, described with hypothetical 1D data.}
    \label{fig:adaptiveSampling}
\end{figure}

For the first iteration of the workflow, there are no data to use to create an IDNN surrogate model for the chemical potential. We instead use the equations for the chemical potentials of an ideal solution, which are the partial derivatives of the ideal solution free energy with respect to the order parameters. With the free energy density and chemical potentials expressed in terms of the more transparent sublattice compositions, the relations are:
\begin{equation}
    \tilde{f}(\bsym{x}) = \frac{k_BT}{4} \sum_{i=1}^4 \left(x_i\log x_i + (1-x_i)\log(1-x_i) \right)
\end{equation}
\begin{equation}
    \widetilde{\mu}_{i-1}(\bsym{x}) := \frac{k_BT}{4} \sum_{j=1}^4 \log\left(\frac{x_j}{1-x_j}\right)Q_{ji}^{-1} \qquad i = 1,..,4 \label{eqn:mu_ideal}
\end{equation}\\
where $k_B$ is the Boltzmann constant and $T$ is the temperature. Figure \ref{fig:adaptiveSampling} and Algorithm \ref{algo:CASMsampling} summarize the full workflow.

\begin{algo}
Active learning of free energy\\
\fbox{\begin{minipage}{11cm}
{\tt
Initialize $k=1$, $\mathcal{D} = \emptyset$, $\widehat{\bsym{\mu}}_0(\bsym{\eta}) = \widetilde{\bsym{\mu}}(\bsym{Q}^{-1}\bsym{\eta})$. Iterate over the following:
\begin{enumerate}
    \item Global sampling:
        \begin{enumerate}
        \item Select sample points in the sublattice composition space:
        \begin{equation*}
            \{\bsym{x} \in (0,1)\times\cdots\times(0,1)\,|\,c \leq 0.25\}
        \end{equation*}
        \item Evaluate the corresponding bias parameter values:
        \begin{equation*}
            \kappa_i = \frac{1}{2\phi_i}\widehat{\mu}_i(\bsym{Q}\bsym{x}) + \sum\limits_j Q_{ij}x_j
        \end{equation*}
        \item Use the $\bsym{\kappa}$ values as input to CASM to compute the order parameter values, $\bsym{\eta}$ and chemical potential values, $\bsym{\mu}$.\\Resulting values form data set $\mathcal{F}_k = \{(\bsym{\eta},\bsym{\mu})\}$.
        \item Update $\mathcal{D} := \mathcal{D} \cup \mathcal{F}_k$.
        \end{enumerate}
    \item Train IDNN surrogate model $\widehat{\bsym{\mu}}_k(\bsym{\eta})$ to the data set $\mathcal{D}$, initialized from  $\widehat{\bsym{\mu}}_{k-1}(\bsym{\eta})$ when $k \ge 2$.
        \begin{enumerate}
        \item Break if $||\widehat{\bsym{\mu}}_k(\bsym{\eta}) - \widehat{\bsym{\mu}}_{k-1}(\bsym{\eta})||_2 < $ tol, \\for $\bsym{\eta}$ sampled using a Sobol$^\prime$ sequence.
        \end{enumerate}
    \item Local (error-based) sampling:
        \begin{enumerate}
        \item Identify points in $\mathcal{F}_k$ that give highest IDNN error.
        \item Identify points with a positive definite Hessian and low gradient norm.
        \item Submit nearby points to CASM; results form data set $\mathcal{L}_k = \{(\bsym{\eta},\bsym{\mu})\}$.
        \item Update $\mathcal{D} := \mathcal{D} \cup \mathcal{L}_k$.
        \item k = k + 1
        \end{enumerate}
\end{enumerate}}
\end{minipage}}
\label{algo:CASMsampling}
\end{algo}

\subsection{FCC symmetry}
\label{sec:fcc_symmetry}
Due to the symmetry of the FCC crystal structure in the Ni-Al system, the free energy density  should be invariant to permutations of $\eta_1,\eta_2,\eta_3$ and changes in the sign of any two of the order parameters $\eta_1$, $\eta_2$, and $\eta_3$ \cite{Braun1997}. To impose this invariance, we express the free energy density as a function of the following invariants:
\begin{equation}
    \begin{split}
        p_1(\eta_1,\eta_2,\eta_3) &= 16\eta_1\eta_2\eta_3\\
        p_2(\eta_1,\eta_2,\eta_3) &= 8(\eta_1^2+\eta_2^2+\eta_3^2)\\
        p_3(\eta_1,\eta_2,\eta_3) &= 4(\eta_1^2\eta_2^2+\eta_2^2\eta_3^2+\eta_3^2\eta_1^2)
    \end{split}
\end{equation}
Thus, the proper symmetry is perfectly enforced by setting $f(c,\eta_1,\eta_2,\eta_3) := \hat{f}(c,p_1,p_2,p_3)$.

\section{Elasticity}
\label{sec:elas}
Strain energy surfaces for fcc Ni and Ni$_{3}$Al were computed from first-principles using density functional theory as implemented in the \emph{Vienna Ab-Initio Simulation Package}. The same input parameter set was used for these calculations as detailed in the study by Goiri and Van der Ven \cite{goiri2016}. Within each phase, 2157 symmetrically distinct homogeneous strains of the equilibrium crystal structure were enumerated using the algorithm described by Thomas and Van der Ven \cite{thomas2017}. For each homogeneous strain value, the atomic coordinates are relaxed to minimize the total energy of the crystal. These strain energy data are provided in the supplementary information.\par
As the strain energy is directly obtained from DFT calculations, a standard DNN can be trained to fit the data. Separate DNNs are trained for pure Ni and for Ni$_3$Al. Similar to the symmetry invariance within the space of order parameters detailed in Sections \ref{sec:first_principles} and \ref{sec:fcc_symmetry}, the strain energy DNN must also be invariant to the symmetries of the FCC crystal. To impose the proper strain invariance associated with cubic symmetry, the strain energy DNN's are trained to be functions of the following symmetry-invariant strain polynomials\cite{thomas2017}:
\begin{equation}
    \begin{split}
        h_1 &= e_1\\
        h_2 &= \sqrt{1/2}(e_2^2 + e_3^2)\\
        h_3 &= \sqrt{1/3}(e_4^2 + e_5^2 + e_6^2)\\
        h_4 &= (1/2)(e_3^3 - 3e_3e_2^2)\\
        h_5 &= e_3(2e_4^2 - e_5^2 - e_6^2)/2 - \sqrt{3}e_2(e_5^2 - e_6^2)/2\\
        h_6 &= \sqrt{6}e_4e_5e_6
    \end{split}
\end{equation}
where $e_i$, $i=1,\ldots,6$ are defined in terms of the elastic part of the Green-Lagrange strain tensor \cite{thomas2017}:
\begin{equation}
    \begin{split}
        e_1 &= (E^e_{11} + E^e_{22} + E^e_{33})/\sqrt{3}\\
        e_2 &= (E^e_{11} - E^e_{22})/\sqrt{2}\\
        e_3 &= (2E^e_{33} - E^e_{11} - E^e_{22})/\sqrt{6}\\
        e_4 &= \sqrt{2}E^e_{23}\\
        e_5 &= \sqrt{2}E^e_{13}\\
        e_6 &= \sqrt{2}E^e_{12}
    \end{split}
\end{equation}
The DNN representation of the strain energy density, $\psi$, and its partial derivatives, $\partial\psi/\partial e_i$, should vanish at zero strain. To enforce this condition, the squares of these terms are added to the loss function to serve as penalties:
\begin{equation}
    \mathrm{loss} := \mathrm{MSE} + \left(\lambda_1\psi^2 + \lambda_2\Big|\frac{\partial\psi}{\partial\bsym{e}}\Big|^2\right)\Bigg|_{\bsym{e}=0}
\end{equation}

The strain energy for the Ni-Al system is dependent on the composition. A linear interpolation of the two strain energy density DNNs is used to model the strain energy at intermediate compositions, as follows:
\begin{align}
    \psi(c,\bsym{F}^e) := (1-4c)\psi^\mathrm{Ni}(\bsym{F}^e) + 4c\psi^\mathrm{Ni_3Al}(\bsym{F}^e)
\end{align}
Note that each strain energy density is written as a function of the elastic part of the deformation gradient, $\bsym{F}^e$.

In addition to the change in strain energy density, the lattice parameter of the Ni-Al system also varies with $c$ \cite{Kamara1996,Arroyave2005}. We represent the function defining the lattice parameter by $a(c)$. For Ni-Al, this relationship between the lattice constant and the composition has been observed to be approximately linear \cite{Kamara1996} and can be modeled by the following function:
\begin{equation}
    a(c) = \frac{a_{\gamma'} - a_\gamma}{c_{\gamma'}-c_\gamma}(c-c_\gamma) + a_\gamma
\end{equation}
where $a_\gamma$ and $a_{\gamma'}$ are the lattice parameters at compositions $c_\gamma$ and $c_{\gamma'}$, respectively (see Table \ref{tab:lattice_pmtr}).
\begin{table}[]
    \centering
    \begin{tabular}{|c | c | c |}
    \hline
    Phase & Composition & Lattice parameter\\
    \hline
        $\gamma$ matrix & 0.12 & 0.356 nm\\
        $\gamma'$ precipitate & 0.23 & 0.358 nm\\
    \hline
    \end{tabular}
    \caption{Lattice parameters of the $\gamma$ matrix and $\gamma'$ precipitate phases used in this work, based on experimental values \cite{Kamara1996}.}
    \label{tab:lattice_pmtr}
\end{table}
The difference in lattice parameter between different phases induces a small misfit strain ($\sim 0.6\%$) in the material. The misfit strain is incorporated through the multiplicative decomposition of the deformation gradient into its elastic and eigenstrain parts, $\bsym{F} = \bsym{F}^e\bsym{F}^\lambda$:

\begin{align}
    \bsym{F}^\lambda &= \lambda\mathbbm{1}\\
    \bsym{F}^e &= \bsym{F}{\bsym{F}^\lambda}^{-1}\\
    \bsym{E}^e &= \frac{1}{2}\left({\bsym{F}^e}^T\bsym{F}^e - \mathbbm{1}\right)
\end{align}
where $\lambda := a(c)/a(\bar{c})$, $\bar{c}$ is the volume-averaged composition, and $\bsym{F}$ is the total deformation gradient.

\section{Phase field formulation}
\label{sec:pf}
We have deployed the analytically integrated free energy DNN and the strain energy DNNs in phase field computations. The phase field model was based on the coupled Cahn-Hilliard and Allen-Cahn equations with quasi-static finite strain elasticity \cite{CahnHilliard1958,Allen1979,Rudrarajuetal2016}.

The composition $c$, which is equal by definition to the order parameter $\eta_0$, is a conserved value. The remaining order parameters $\eta_1, \eta_2, \eta_3$ are nonconserved variables. Given the homogeneous free energy density $f(c,\eta_1, \eta_2, \eta_3)$ as a function of composition and order parameters, we define the total free energy as the following:
\begin{equation}
\begin{split}
    \Pi[c,\eta_1, \eta_2, \eta_3, \bsym{u}] = \int\limits_\Omega \Big[f(c,\eta_1, \eta_2, \eta_3) + \psi(c,\bsym{F}^e) + \frac{1}{2}\chi_0|\nabla c|^2 + \sum_{i=1}^3\frac{1}{2}\chi_i|\nabla\eta_i|^2\Big]\,\mathrm{d}V
\end{split}
\end{equation}
The corresponding chemical potentials are obtained by computing the variational derivatives of the total free energy, namely $\delta\Pi/\delta c$ and $\delta\Pi/\delta\eta_i$, $i = 1,2,3$. Using standard variational calculus results in the following equations for the chemical potentials:
\begin{align}
    \mu_0 &= \frac{\partial f}{\partial c} + \frac{\partial \psi}{\partial c} + \bsym{P}:\frac{\partial \bsym{F}^e}{\partial c} - \chi_0 \nabla^2 c \label{eqn:mu_0}\\
    \mu_i &= \frac{\partial f}{\partial \eta_i} - \chi_i \nabla^2 \eta_i, \qquad i=1,2,3
\end{align}
where $\bsym{P}:=\partial\psi/\partial\bsym{F}^e$ is the first Piola-Kirchhoff stress tensor.

The phase field model consists of the Cahn-Hilliard and Allen-Cahn equations, given by the following, respectively:
\begin{align}
    \frac{\partial c}{\partial t} &= -\nabla\cdot\bsym{J} \label{eqn:CH}\\
    \frac{\partial \eta_i}{\partial t} &= -L\mu_i, \qquad i=1,2,3
\end{align}
where $L$ is the kinetic coefficient.
The Cahn-Hilliard equation is in conservation form, with the flux defined as $\bsym{J} := -M\nabla\mu_0$, where $M$ is the mobility. It models the overall composition of the system through $c$, while conserving mass. The Allen-Cahn equations model the time evolution of the long-range ordering of the system through the non-conserved order parameters $\eta_1, \eta_2, \eta_3$. The equations are coupled since the chemical potentials are derived from the same free energy. Since elastic equilibrium is attained much more rapidly than phase evolution, the corresponding governing equations of finite strain (nonlinear) elasticity are used in the model.

When Eq. (\ref{eqn:mu_0}) is substituted into Eq. (\ref{eqn:CH}), using the constitutive relation $\bsym{J} := -M\nabla\mu_0$, the result is a fourth-order differential equation. The weak form of this PDE can be solved directly using isogeometric analysis \cite{CottrellHughesBazilevs2009}, due to the higher order continuity of NURBS. However, to solve the equation using the finite element method, we employ a mixed formulation that expresses the fourth order PDE as two second order PDEs. In this formulation, $c$ and $\mu_0$ are both (coupled) primal fields, in addition to the order parameters $\eta_1, \eta_2, \eta_3$ and the displacement field $\bsym{u}$. Thus, the phase field model is described using the following set of second order PDEs in strong form, coupled through the free energy and strain energy density functions:

\begin{align}
    \text{\bf Cahn-Hilliard: }& &\frac{\partial c}{\partial t} &= \nabla\cdot\left(M\nabla\mu_0\right)\\
    & &\mu_0 &= \frac{\partial f}{\partial c} + \frac{\partial \psi}{\partial c} + \bsym{P}:\frac{\partial \bsym{F}^e}{\partial c} - \chi_0 \nabla^2 c\\
    \text{\bf Allen-Cahn: }& &\frac{\partial \eta_i}{\partial t} &= -L\left(\frac{\partial f}{\partial \eta_i} - \chi_i \nabla^2 \eta_i\right),  \qquad i=1,2,3\\
    \text{\bf Elasticity: }& &0 &=\nabla\cdot\left(\bsym{P}{\bsym{F}^\lambda}^{-T}\right)
\end{align}

For the equations written as above, the following Neumann boundary conditions are applied to $c$, $\eta_1$, $\eta_2$, $\eta_3$ and $\mu_0$,  on $\partial\Omega$, where $\bsym{n}$ is the outward unit normal:\footnote{See Ref \cite{Rudrarajuetal2016} for a variational treatment of the boundary conditions on the Cahn-Hilliard equations.}
\begin{align}
    \nabla c\cdot\bsym{n} &= 0\label{eq:dirbc-c}\\
    \nabla \eta_i\cdot\bsym{n} &= 0, \qquad i=1,2,3\label{eq:dirbc-eta}\\
    \nabla \mu_0\cdot\bsym{n} &= 0\label{eq:neumbc-mu0}
\end{align}
A homogeneous Dirichlet boundary condition is applied to the displacement field on $\partial\Omega$:
\begin{align}
    \bsym{u} &= \bsym{0}
\end{align}

The above treatment of Cahn-Hilliard and Allen-Cahn phase field models coupled with elasticity builds on the work of other authors \cite{thornton12004,thornton22004,Wang2019,LQChen2013,LQChen2014}. The corresponding infinite dimensional weak form of the equations for the case with a uniform mobility, as solved using a mixed finite element method, is the following:
\begin{align}
    0 &= \int_\Omega \left(w_c\frac{\partial c}{\partial t} + M\nabla w_c\cdot\nabla\mu_0\right)\mathrm{d}V\\
    0 &= \int_\Omega \left[w_{\mu_0}\left(\mu_0 - \frac{\partial f}{\partial c} - \frac{\partial \psi}{\partial c} - \bsym{P}:\frac{\partial \bsym{F}^e}{\partial c}\right) - \chi_0\nabla w_{\mu_0}\cdot\nabla c\right]\mathrm{d}V\\
    0 &= \int_\Omega \left[w_{\eta_i}\frac{\partial \eta_i}{\partial t} + L\left(w_{\eta_i}\frac{\partial f}{\partial\eta_i} + \chi_i\nabla w_{\eta_i}\cdot\nabla\eta_i\right)\right]\mathrm{d}V, \qquad i = 1,2,3\\
  0 &=  \int_\Omega \nabla \bsym{w}_u:\left(\bsym{P}{\bsym{F}^\lambda}^{-T}\right)\,\mathrm{d}V
\end{align}
where $w_c,w_{\mu_0},w_{\eta_i}$ and $\bsym{w}_u$ are weighting functions.

\section{Implementation and results}
\label{sec:results}
The workflow described in Section \ref{sec:active} was run on the ConFlux high performance computing cluster at the University of Michigan, with the \texttt{CASM} Monte Carlo runs taking place on the CPU nodes and training of the IDNN, implemented with \texttt{Keras} and \texttt{Tensorflow}, utilizing GPUs. Between one and two thousand new points were calculated with each global sampling, and up to 2,800 new data points were added with each local sampling. Over 58,000 data points had been sampled by the end of the 16th iteration of the workflow in Algorithm 1. These data are provided in the supplementary information. The values of the chemical potentials were temporarily scaled by 100$\times$ to improve the training of the IDNN.

\begin{figure}[tb]
        \centering
\begin{minipage}[t]{0.5\textwidth}
        \centering
	\includegraphics[width=\textwidth]{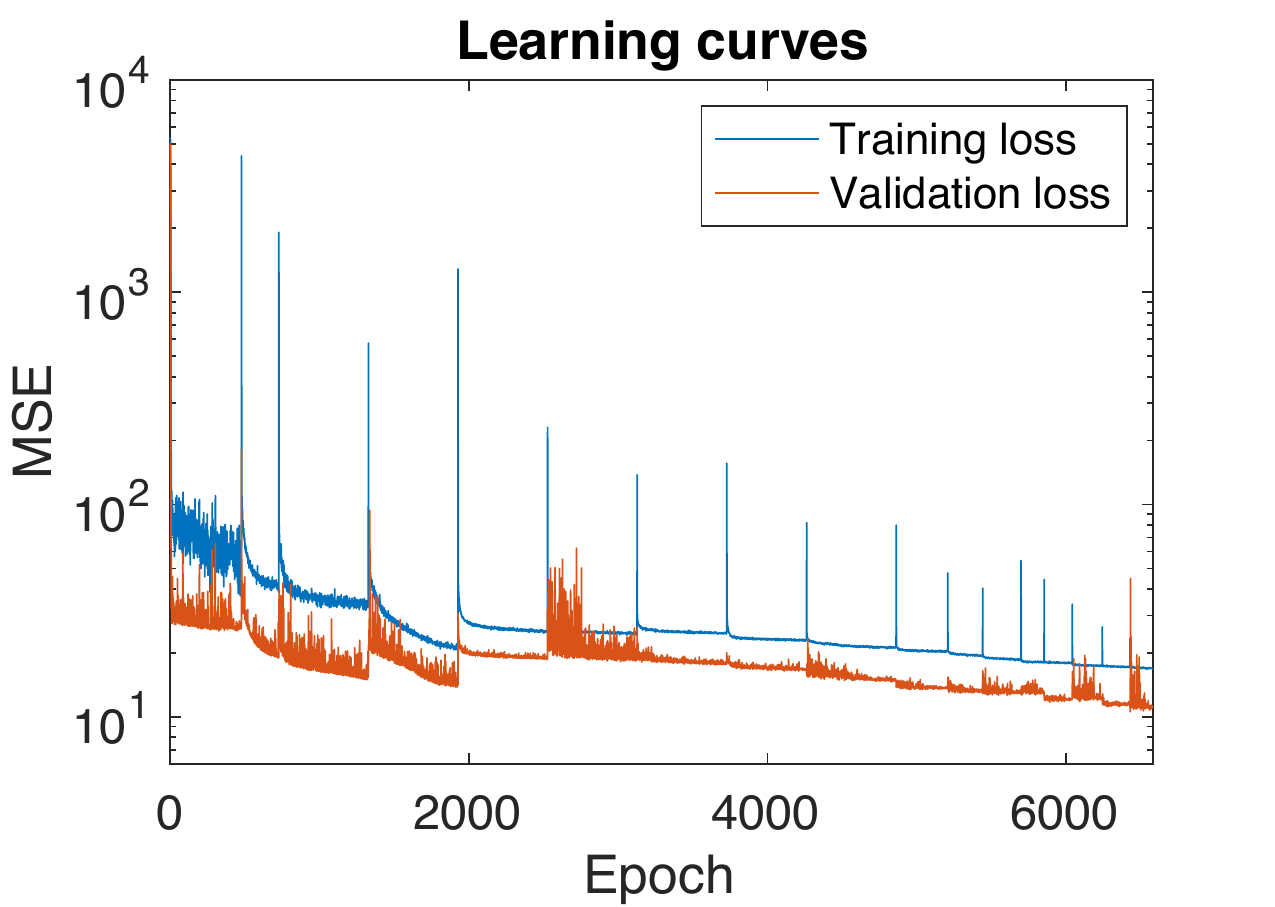}
	\subcaption{}
\end{minipage}%
\begin{minipage}[t]{0.5\textwidth}
        \centering
	\includegraphics[width=\textwidth]{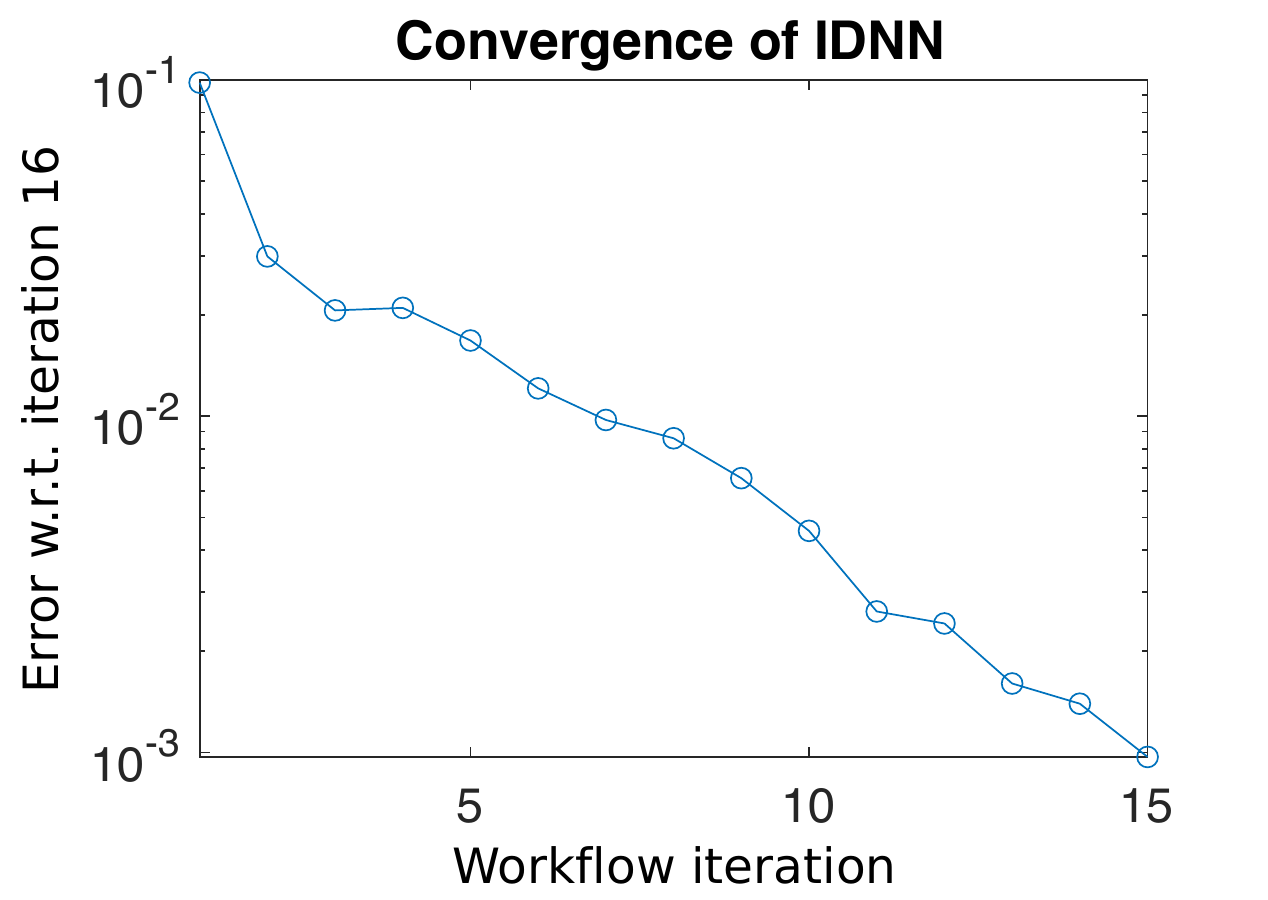}
	\subcaption{}
\end{minipage}
\caption{(a) Learning curves for the IDNN training over the active learning workflow. Periodic jumps in the loss occur at the beginning of each new iteration of the workflow. (b) Convergence of the IDNN over iterations of the active learning workflow is shown by computing the L$_2$-norm of the difference in predicted chemical potential values between each iteration and the final iteration, i.e. $||\bsym{\mu}_\mathrm{final} - \bsym{\mu}_i||_2$. The workflow converged with a tolerance of $1\times10^{-3}$ by the 16th iteration, with $||\bsym{\mu}_{16} - \bsym{\mu}_{15}||_2 = 9.7\times10^{-4}$.}
\label{fig:convergence}
\end{figure}

Since the data after the first global sampling were still quite sparse, a hyperparameter search was performed only after the second global sampling. The IDNN in the first iteration of the workflow was set to have two hidden layers with 20 units each and a learning rate of 0.2. The hyperparameter search was performed by comparing 22 IDNN architectures and learning rates. Learning rate values were randomly chosen log-uniformly from the domain $[0.005,0.5]$, and the units per layer were chosen uniformly from the domain $[20,200]$. We kept the number of hidden layers low for two reasons. First, the IDNN will be evaluated at every quadrature point in the phase field simulation, so it is beneficial to have a small network to reduce computation time. As such, complex architectures were not considered, even though any differentiable neural network is capable of being the parent architecture for an IDNN. Second, while the combination of softplus and sigmoid activation functions works well with the IDNN structure, sigmoid activation functions suffer from the vanishing gradient pathology during training if they are very deep \cite{Glorot2010,Zheng2015}. Thus, all IDNNs were set with two hidden layers. A dropout rate of 0.06 was used for both hidden layers. The dropout rate was manually tuned to discourage spurious wells without altering the form of the true energy wells. Each of the 22 IDNNs was trained for 250 epochs, and the IDNN with the lowest validation loss was chosen. With this approach, an initial learning rate of 0.415 and two hidden layers with a width of 84 units were selected. For all subsequent iterations of the workflow, the architecture of the IDNN was kept fixed, and the training of the weights and biases resumed at each new workflow iteration without reinitialization. Additional details concerning the neural network architecture are included in \ref{app:idnn_arch}.

The learning curves for the full workflow are shown in Figure \ref{fig:convergence}a. The IDNN was trained for 600 epochs in each workflow iteration using the \texttt{AdagradOptimizer}. Training was terminated early for a workflow iteration if there was no decrease in the validation loss for 150 consecutive epochs. A learning rate decay of 0.9 was multiplied at each new iteration of the workflow. Additionally, the learning rate was temporarily reduced by half whenever the validation loss plateaued for 100 epochs, then reset at the beginning of the next workflow iteration. Periodic jumps in the loss occur at the beginning of each new iteration of the workflow, as new data are added to the set. For the first six iterations, the search space is slightly expanded to oversample the edges of the physical domain and resolve the data as the chemical potentials diverge according to the ideal solution equation (\ref{eqn:mu_ideal}).

Convergence of the IDNN is shown by computing the L$_2$-norm of the difference in predicted chemical potential values between each iteration and the final iteration, i.e. $||\bsym{\mu}_\mathrm{final} - \bsym{\mu}_i||_2$. Each IDNN is evaluated at $c,\eta_1,\eta_2,\eta_3$ values determined by a Sobol$^\prime$ sequence to approximate the integration in the L$_2$-norm \cite{Sobol1967}. The workflow converged with a tolerance of $1\times10^{-3}$ by the 16th iteration, with $||\bsym{\mu}_{16} - \bsym{\mu}_{15}||_2 = 9.7\times10^{-4}$, as seen in Figure \ref{fig:convergence}b.

\begin{figure}
    \centering
    \includegraphics[width=\textwidth]{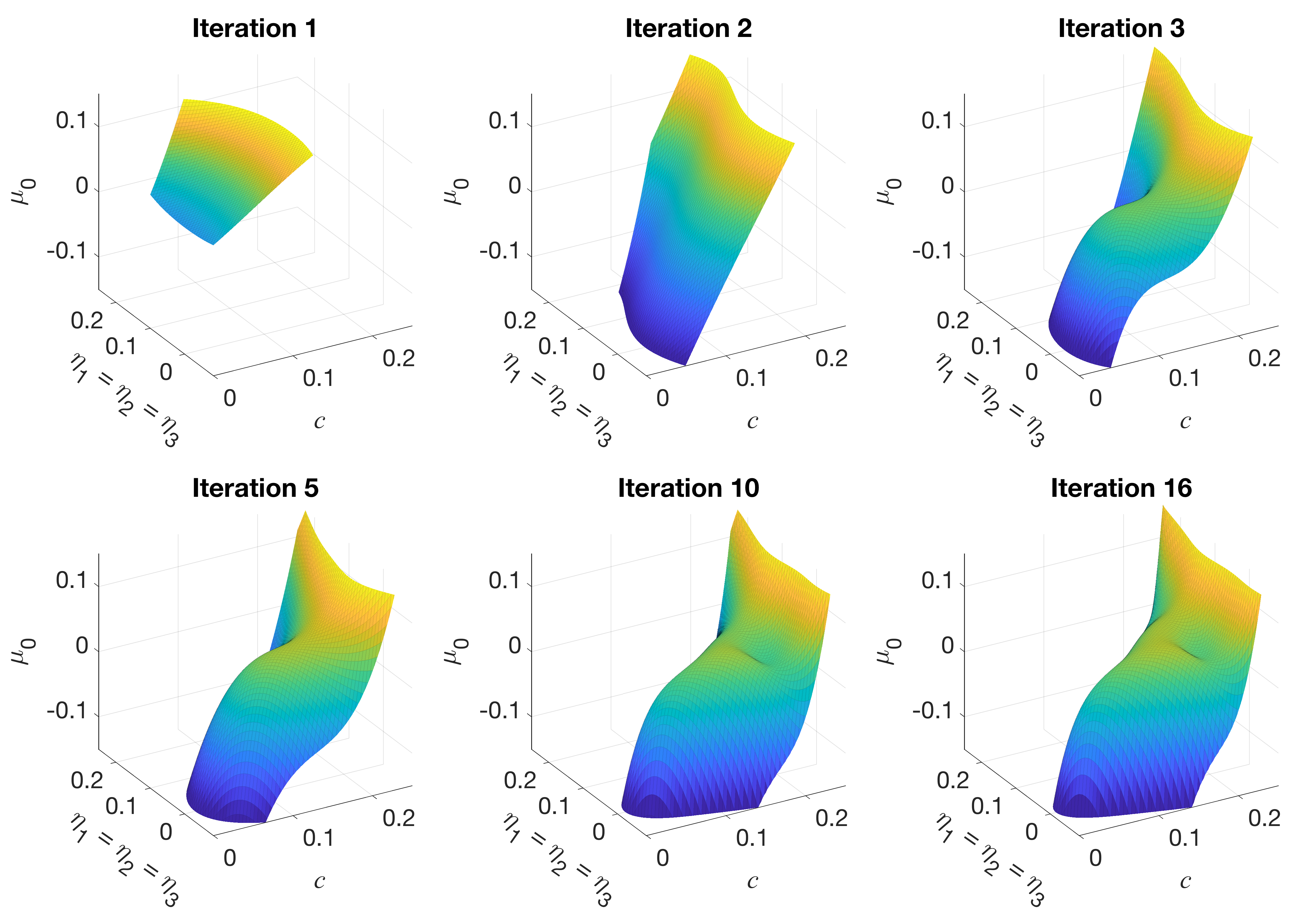}
    \caption{Evolution of the IDNN representing the chemical potential for a two-dimensional subspace, over iterations of the active learning workflow.}
    \label{fig:mu_plots}
\end{figure}

\begin{figure}
    \centering
    \includegraphics[width=\textwidth]{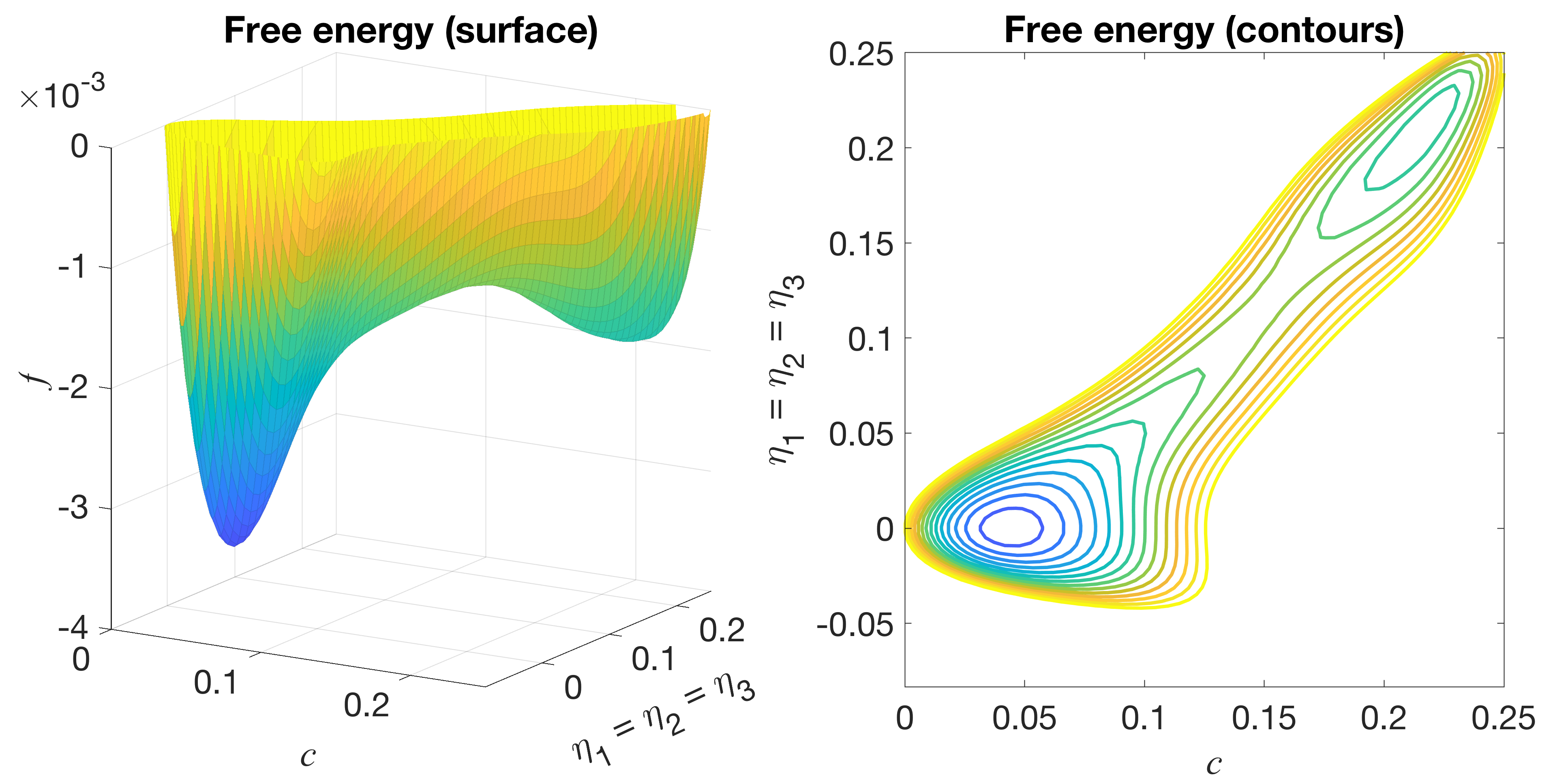}
    \caption{Surface and contour plots for a 2D subspace ($\eta_1 = \eta_2 = \eta_3$) of the converged DNN representation of the homogeneous free energy density.}
    \label{fig:free_en}
\end{figure}

\begin{figure}[tb]
        \centering
\begin{minipage}[t]{0.45\textwidth}
        \centering
	\includegraphics[width=0.9\textwidth]{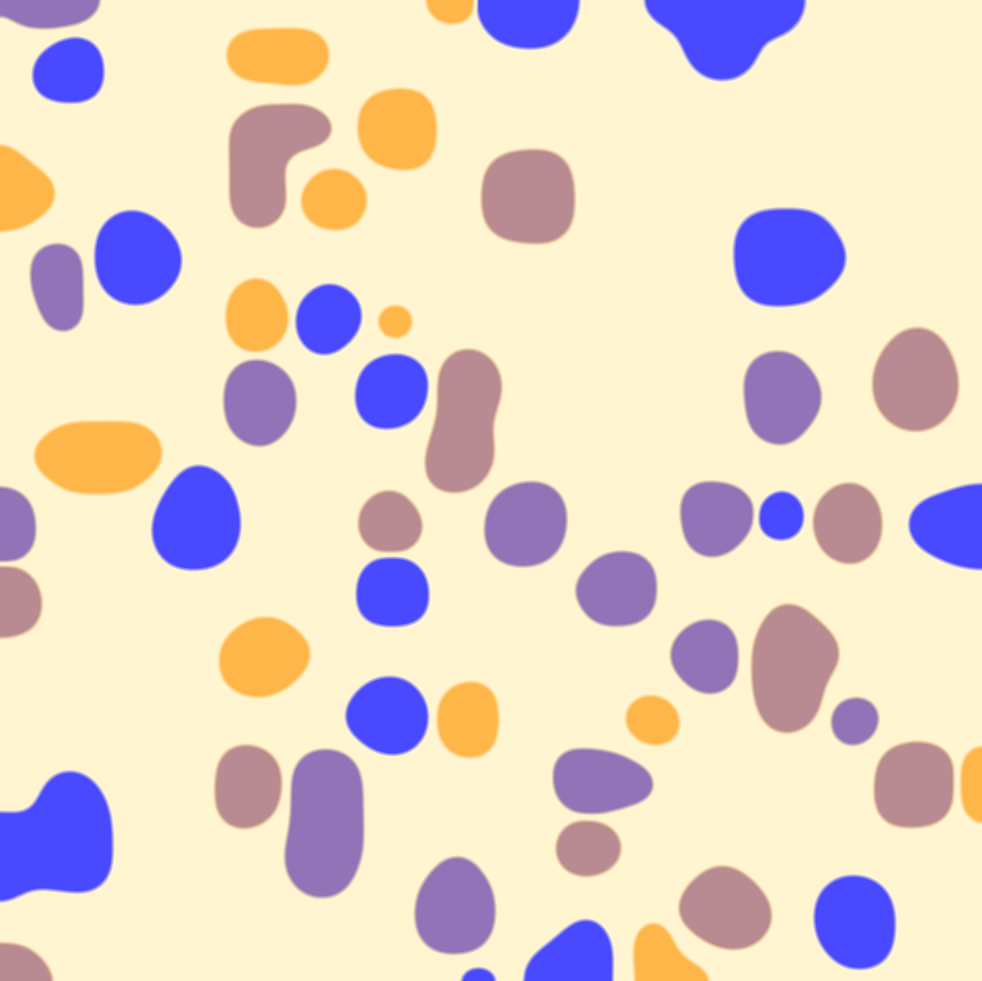}
	\subcaption{Time step 50}
\end{minipage}%
\begin{minipage}[t]{0.45\textwidth}
        \centering
	\includegraphics[width=0.9\textwidth]{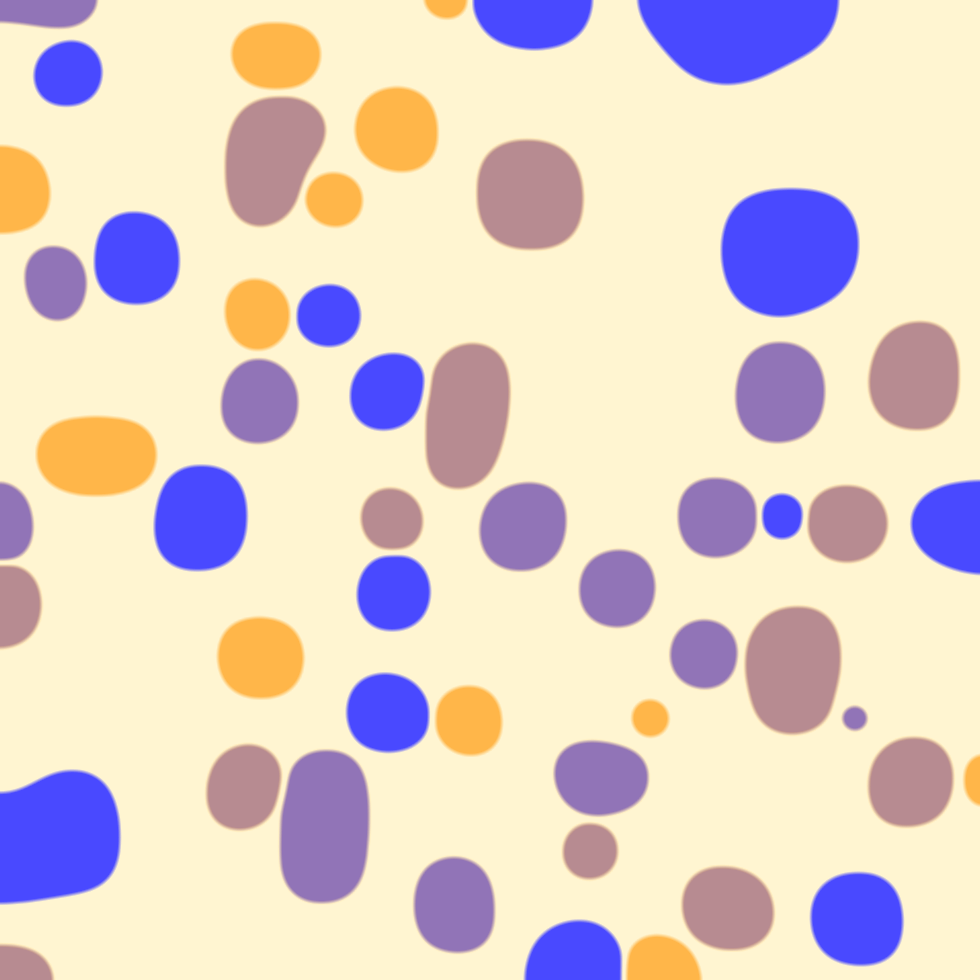}
	\subcaption{Time step 150}
\end{minipage}
\begin{minipage}[t]{0.45\textwidth}
        \centering
	\includegraphics[width=0.90\textwidth]{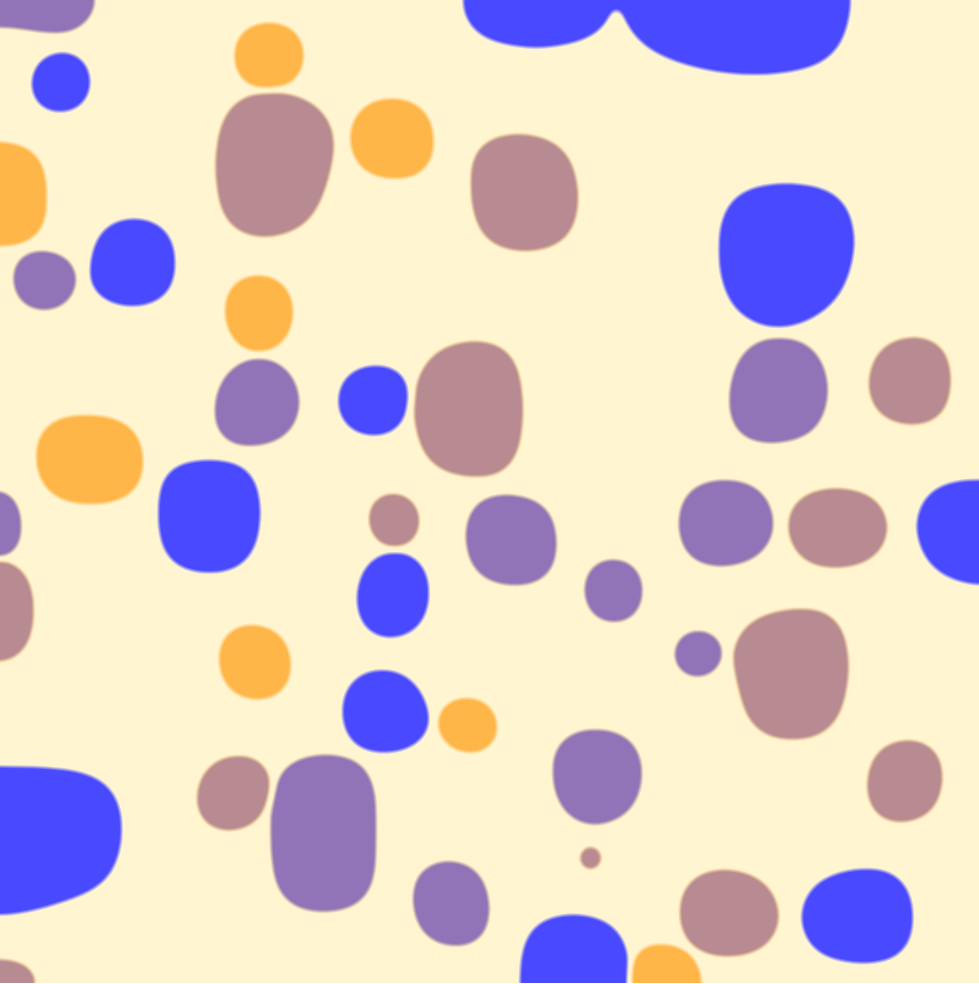}
	\subcaption{Time step 300}
\end{minipage}%
\begin{minipage}[t]{0.45\textwidth}
        \centering
	\includegraphics[width=0.90\textwidth]{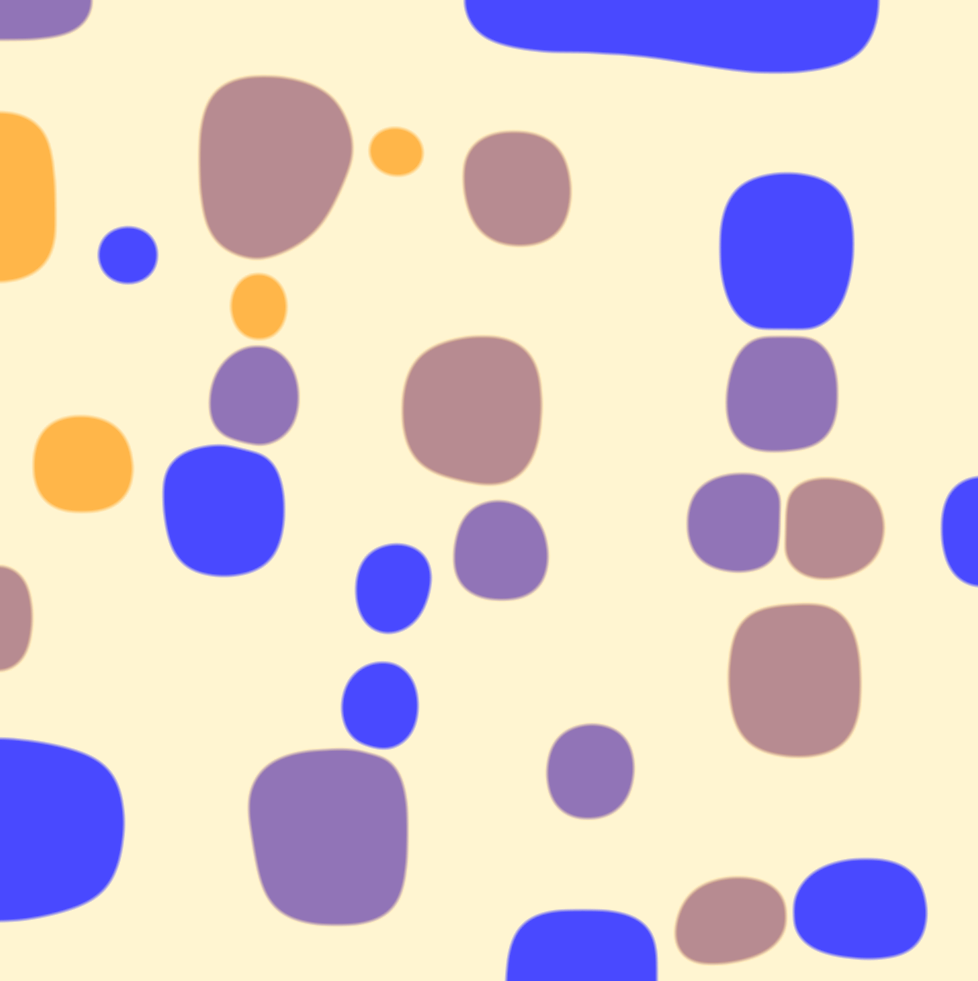}
	\subcaption{Time step 650}
\end{minipage}
\caption{Evolution of the Ni$_3$Al precipitates in the phase field simulations, with four L1$_2$ variants shown as blue, orange, brown, and purple.}
\label{fig:dnn_pf}
\end{figure}

The evolution of the IDNN is presented in Figure \ref{fig:mu_plots} by plotting a slice of the predicted chemical potential $\mu_0$ as a function of $c$ and $\eta_1$, with $\eta_1 = \eta_2 = \eta_3$. Significant changes are seen in the first few iterations of the workflow, with evident convergence in the later iterations. A slice of the final, analytically integrated free energy DNN, referenced to pure Ni and the perfectly ordered L1$_2$, is shown in Figure \ref{fig:free_en}, again with $\eta_1 = \eta_2 = \eta_3$. An energy well is seen at about $c = 0.23$, corresponding to the $\gamma^\prime$ Ni-Al precipitates for the L1$_2$ variant with all positive valued order parameters. A well near $c = 0.045$ represents the $\gamma$ solid solution phase. A few spurious regions of slight convexity exist in the DNN surface, but they do not seem to negatively affect the resulting precipitate formation in the phase field results (see Figure \ref{fig:dnn_pf}).

\begin{figure}
    \centering
    \includegraphics[width=0.5\textwidth]{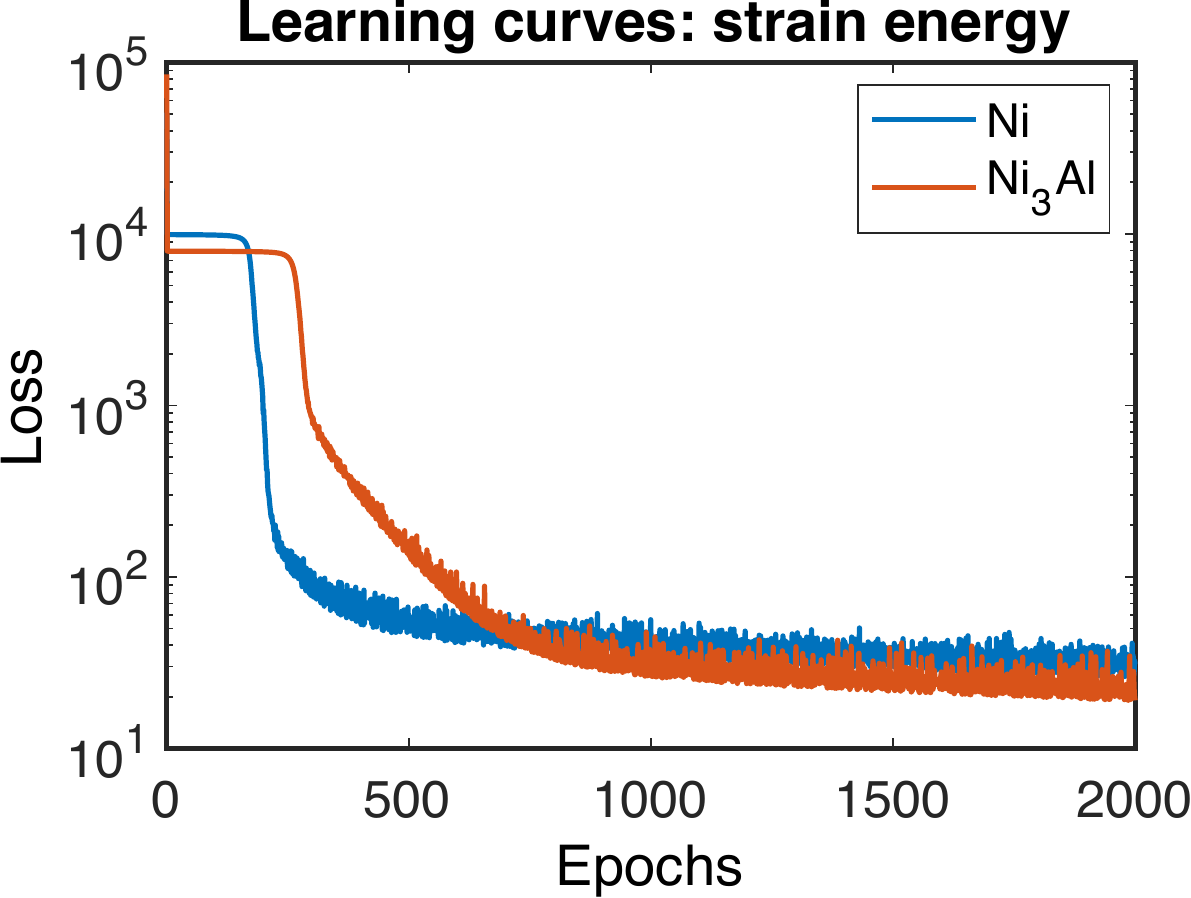}
    \caption{Training loss for the Ni and Ni$_3$Al strain energy density DNNs.}
    \label{fig:lc_strain}
\end{figure}

\begin{figure}
    \centering
    \includegraphics[width=0.8\textwidth]{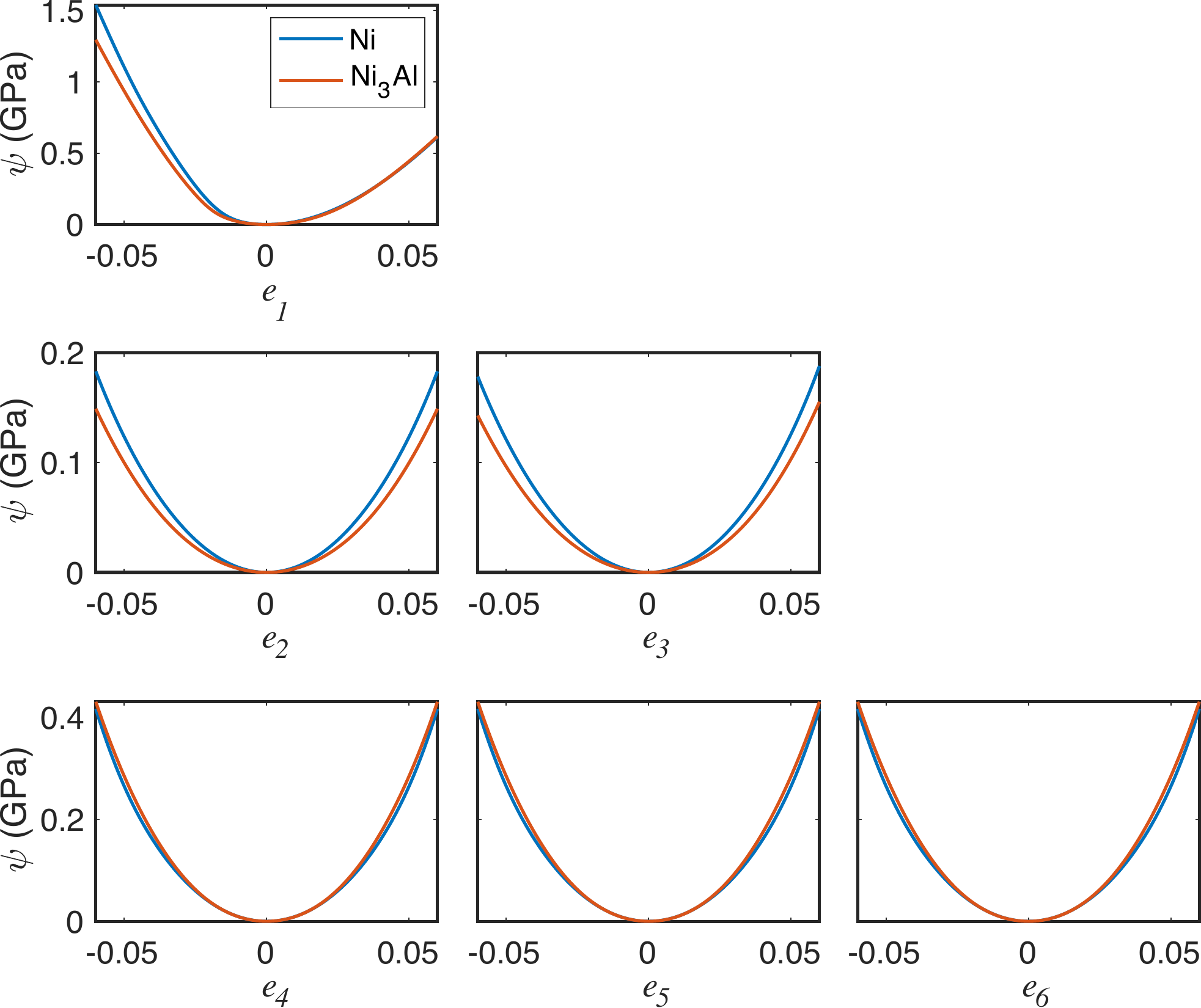}
    \caption{One-dimensional slices of the Ni and Ni$_3$Al strain energy density DNNs, plotted against each of the strain metrics $e_i$, $i=1,\ldots,6$.}
    \label{fig:strain_energy}
\end{figure}

The strain energy density data were used to train two DNNs while enforcing appropriate symmetries, as described in Section \ref{sec:elas}. The DNNs were each defined with two hidden layers of 60 activation units each, with the softplus activation function for smoothness. The strain energy values were temporarily scaled by 100$\times$ to improve the training of the DNNs, and each DNN was trained for 2000 epochs (see the learning curves in Figure \ref{fig:lc_strain}). Because of the penalty term in the loss function, the two strain energy DNNs predicted an energy very close to zero with an input of zero. The bias on the output layer of each DNN was further modified to exactly enforce the condition of zero energy at zero strain. One-dimensional slices of the strain energy density are plotted against each of the strain metrics $e_i$, $i=1,\ldots,6$ in Figure \ref{fig:strain_energy}.

The phase field equations from Section \ref{sec:pf} were solved numerically using the finite element method and backward Euler time integration. The simulation was performed using the \texttt{mechanoChemFEM} code\footnote{Code available at github.com/mechanoChem/mechanoChemFEM}, which is based on the \texttt{deal.II} \cite{dealii2017} library, and run on the ConFlux HPC cluster at the University of Michigan. Initial conditions were random about $c = 0.1$ and $\eta_i = 0$, $i = 1,2,3$. Adaptive time stepping and adaptive mesh refinement were used. Values of $\chi_0 = 1.224\times10^{-3}$ and $\chi_i = 4.9\times10^{-3}$, $i=1,2,3$ were used for the interfacial and anti-phase gradient parameters, respectively. These result in an interfacial energy of about 45 mJ/m$^2$, which is the correct order of magnitude for Ni-Al \cite{ardell1966,mishin2004}.

The simulation results are plotted in Figure \ref{fig:dnn_pf}. A movie of the simulation results is also provided in the supplementary information. The expected development of precipitates of multiple L1$_2$ variants are shown, demonstrating that the free energy DNN has captured the appropriate physics. The blue, orange, brown, and purple regions represent the four $L1_2$ variants that develop in the Ni-Al precipitates at about $c = 0.23$. The tan background shows the $\gamma$ phase solid solution. As expected, each precipitate consists of a single L$1_2$ variant, without any antiphase boundaries forming within a single precipitate. This is due to the antiphase boundary energy being greater than twice the interfacial energy, as described by Wang, et al \cite{wang1998}. The precipitates are seen to grow over time, forming roughly rectangular shapes. This faceting is due to the interplay between the misfit strain, interfacial energy, and cubic elasticity. Note that the smaller precipitates remain circular, since the interfacial energy is dominating the anisotropic elastic response \cite{Voorhees2004}.

\section{Conclusions}
\label{sec:conclusions}

This communication advances the treatment of scale bridging for the thermodynamics relevant to a real system in materials physics. Starting with electronic structure computations, we have ascended the scales through statistical mechanics descriptions to the PDEs of continuum physics. Notably, this treatment has combined the mechano-chemical interactions that reside in free energy functions, avoiding phenomenology. This advance in scale bridging for materials physics has leveraged several separate innovations in machine learning, and more broadly, in data-driven modelling.

We have developed an active learning workflow to improve sampling of chemical potential data while simultaneously constructing a deep neural network (DNN) representation of the free energy. The application of active learning, in which the machine learning method identifies regions of the space where more data are needed, also drives the sampling of the high-dimensional space in those regions. 

Using an integrable deep neural network (IDNN) to train to the chemical potential provides an analytically integrated free energy density DNN. This integrability is critical in mechano-chemical coupling, wherein the stresses are defined as derivatives of the free energy density with respect to strains. However, even in the absence of coupling to elasticity, it is essential to maintain consistency of the free energy/chemical potential representation. In this context, na\"{i}vely training to derivative data without enforcing consistency by ensuring a unique antiderivative (up to constants) will manifest as unphysical results: The chemical potentials will not reflect the appropriate physics inherent in their being derivatives of a single free energy density function. In this work, strain energy DNNs were trained separately to strain energy data for Ni and Ni$_3$Al.

To demonstrate that the resulting free energy DNN accurately reflects the physics of the Ni-Al system, we performed phase field simulations using the integrated free energy DNN and the strain energy DNNs as input. The 2D phase field results show the creation, growth, and coarsening of quasi-rectangular Ni$_3$Al precipitates with four variants of anti-phase domains. 

The systematic extraction of the thermodynamic description, originating in first principles computations has culminated in a ten-dimensional free energy function that reflects at least some features of mechano-chemical coupling \cite{Rudrarajuetal2016}.  We recognize that some aspects of our treatment remain phenomenological: The strain energy function has used the rule of mixtures to interpolate between DFT data-based representations for Ni and Ni$_3$Al, rather than attempt a full parameterization of the coupled composition-strain space. However, given that the lattice parameters have only a linear dependence on composition, and the similarity of the Ni and Ni$_3$Al strain energy representations in Figure \ref{fig:strain_energy}, this is expected to be a weak effect. The gradient parameters in the phase field models were chosen to approximate the interfacial energy on the basis of a simplified, one-dimensional estimate \cite{Teichert2018a}, which, nevertheless remains the prevailing approach. Finally, kinetic parameters of the Ni-Al system also could be obtained from first principles and statistical mechanics \cite{goiri2019}, and machine learning representations developed for them. The ordering process is expected to equilibrate rapidly as it involves only local rearrangements of atoms. The time scale of the microstructure evolution is therefore most likely dominated by long-range diffusion processes as described by the Cahn-Hilliard evolution equation. Building on our previous work \cite{Teichert2018a,Teichert2019}, the results here continue to demonstrate the effectiveness of machine learning methods in enhancing the predictive capabilities of computational physics.


\section{Acknowledgements}
\label{sec:acknowledgements}
We gratefully acknowledge the support of Toyota Research Institute, Award \#849910, ``Computational framework for data-driven, predictive, multi-scale and multi-physics modeling of battery materials''. This work has also been supported in part by National Science Foundation DMREF grant \#1729166, ``Integrated Framework for Design of Alloy-Oxide Structures''. Additional support was provided by Defense Advanced Research Projects Agency (DARPA) under Agreement No. HR0011199002, ``Artificial Intelligence guided multi-scale multi-physics framework for discovering complex emergent materials phenomena''. Simulations in this work were performed using resources provided by the NSF via grant 1531752 MRI: Acquisition of Conflux, A Novel Platform for Data-Driven Computational Physics (Tech. Monitor: Ed Walker), with past and current support by the University of Michigan. Additional computing resources were provided by the Extreme Science and Engineering Discovery Environment (XSEDE) Comet at the San Diego Supercomputer Center through allocation TG-DMR180072. XSEDE is supported by National Science Foundation grant number ACI-1548562.



\clearpage
\appendix
\section{IDNN architecture and optimizer comparisons}
\label{app:idnn_arch}
Several other neural network parameters could have been used in the IDNN used in this work. Here, we present a small exploration of a few of these parameters, namely, the choices of activation function, weight initializer, and optimizer. If desired, each of these neural network parameters could be included in the hyperparameter search. For this comparison, each parameter choice was considered by training 10 IDNNs, with random initializations, for 50 epochs using the Ni-Al chemical potential dataset from the 16th iteration of the active learning workflow. We used the neural network architecture and learning rate chosen from the hyperparameter search, which were two hidden layers with 84 neurons per layer and a learning rate of 0.415. In each case, the resulting learning curves were smoothed using a moving average with a five epoch window for clarity.

\begin{figure}
    \centering
    \includegraphics[width=0.90\textwidth]{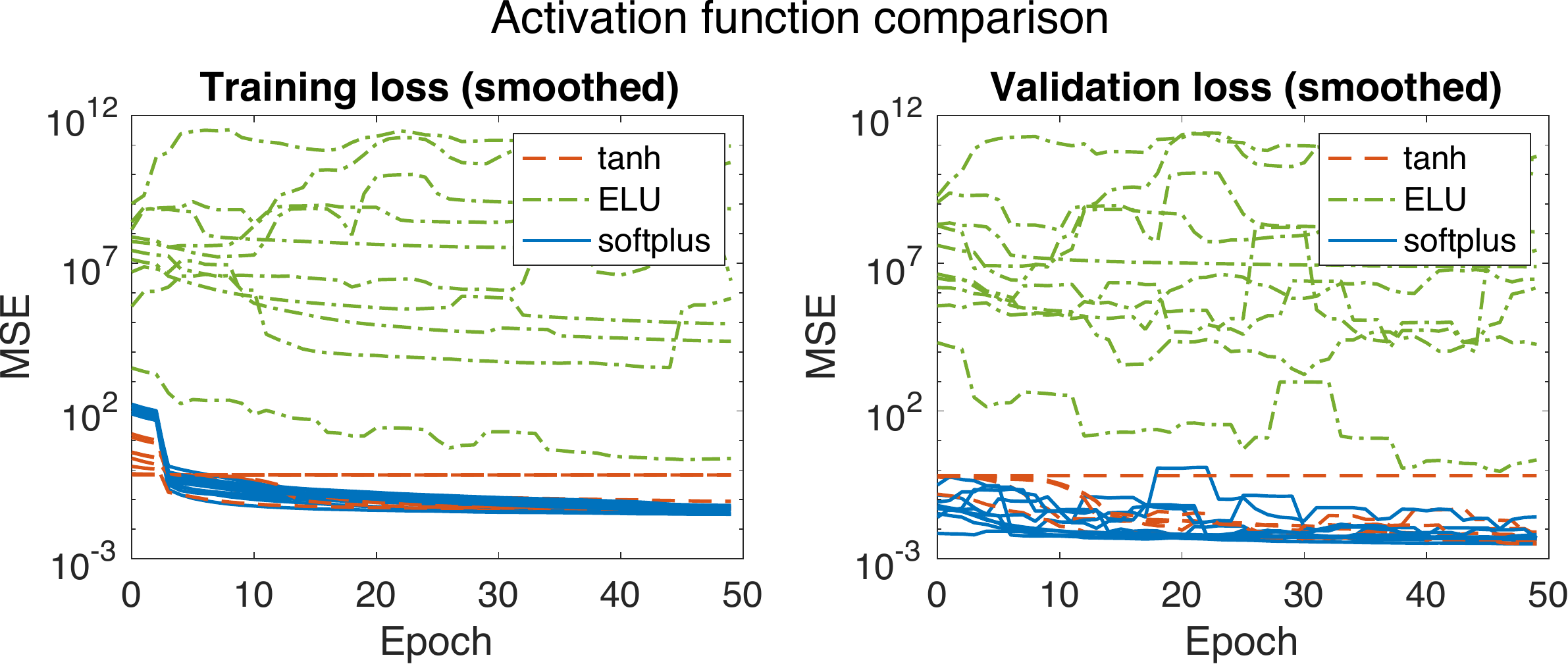}
    \caption{Comparison of learning curves for IDNNs using softplus, ELU, and tanh activation functions, with ten curves of each type.}
    \label{fig:act_fun}
\end{figure}

As described in Section \ref{sec:IDNN}, the combination of softplus and sigmoid activation functions is a logical choice for use in the IDNN because they are the most common pair of activation functions where one is the derivative of the other. It is possible, however, to use other activation functions. In Figure \ref{fig:act_fun}, the training of an IDNN with the softplus function, as used in this work, is compared with IDNNs using the ELU (exponential linear unit) and hyperbolic tangent functions. The softplus activation function gives the lowest values overall for both training and test loss, but the tanh function might also be a reasonable choice. The ELU function does not perform well, perhaps due to its piecewise definition.

\begin{figure}
    \centering
    \includegraphics[width=0.90\textwidth]{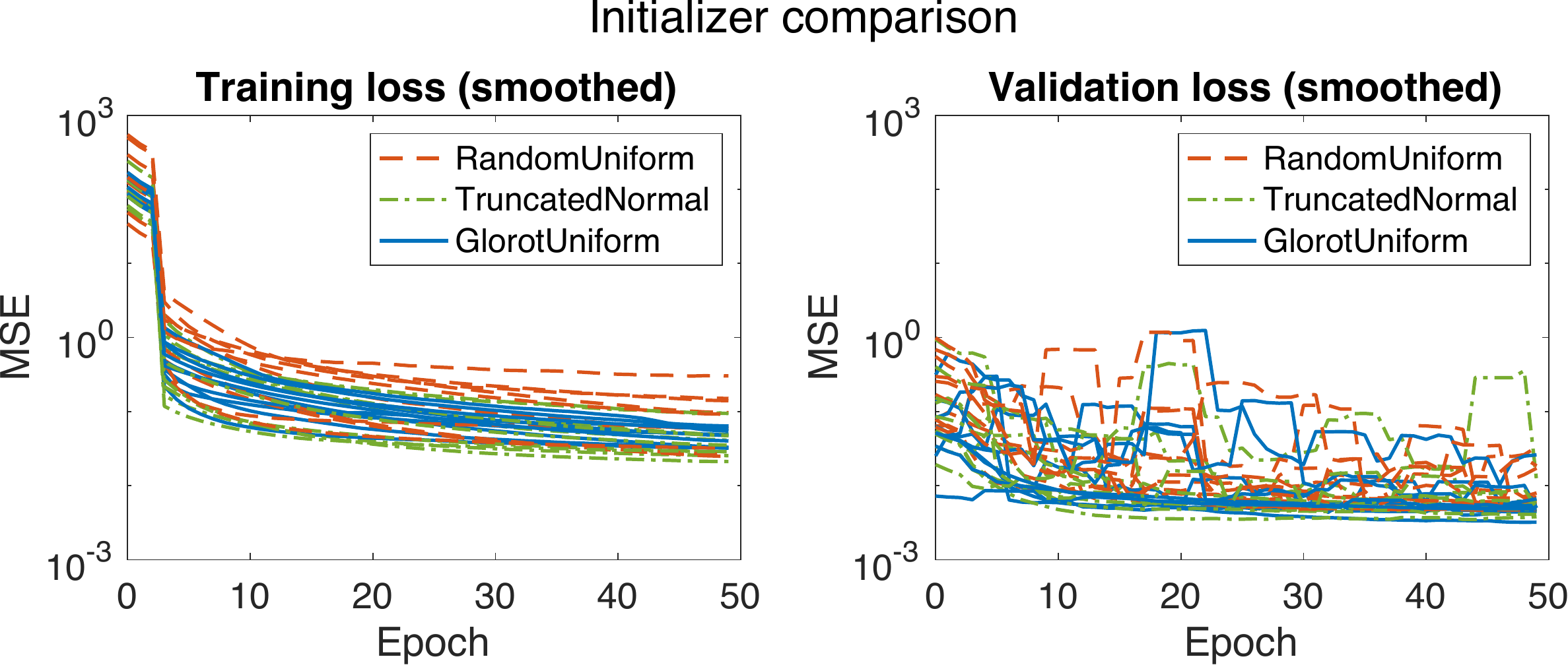}
    \caption{Comparison of learning curves for IDNNs using Glorot uniform, truncated normal, and random uniform initializers, with ten curves of each type.}
    \label{fig:init}
\end{figure}

The results presented in the main body of work used the Glorot uniform initializer for the weights, which is the default weight initializer defined in the \texttt{Keras} library. Other initializers can be used. Figure \ref{fig:init} compares the Glorot uniform initializer with the truncated normal and random uniform intializers. Since the trends for all three initializers are very similar, any of the choices could likely be used without a significant impact on the results.

\begin{figure}[tb]
        \centering
\begin{minipage}[t]{\textwidth}
        \centering
	\includegraphics[width=0.90\textwidth]{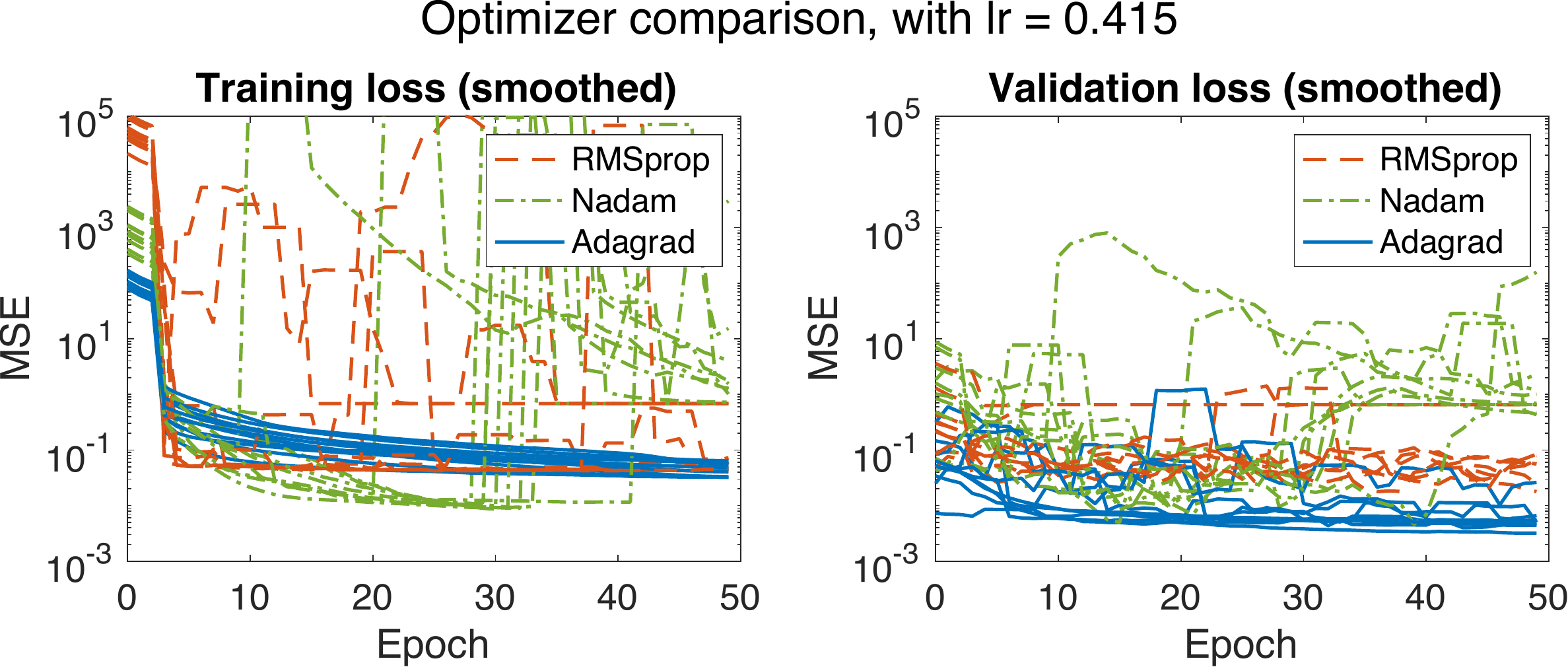}
	\subcaption{}
	\vspace{0.5cm}
\end{minipage}
\begin{minipage}[t]{\textwidth}
        \centering
	\includegraphics[width=0.90\textwidth]{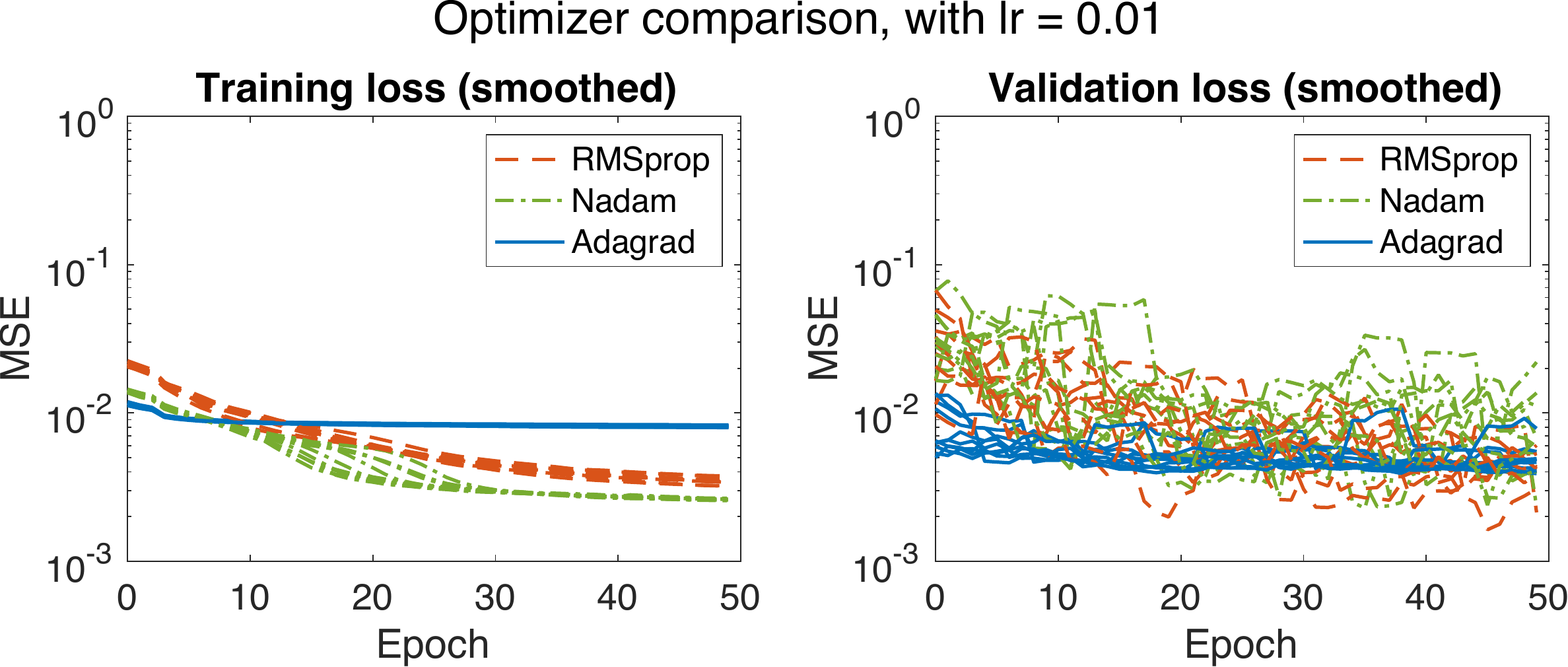}
	\subcaption{}
\end{minipage}
\caption{Comparison of learning curves for IDNNs using RMSprop, Nadam, and Adagrad optimizers, with ten curves of each type. Results using two different learning rates are reported: (a) learning rate = 0.415, and (b) learning rate = 0.01.}
\label{fig:opt}
\end{figure}

Several different variations of gradient descent have been developed and used in deep learning. As stated in Section \ref{sec:results}, we used the Adagrad optimizer in this work. The Adagrad optimizer is compared with results using the  RMSprop and Nadam optimizers in Figure \ref{fig:opt}a. By the 50th epoch, Adagrad had, in most cases, the lowest overall error. Adagrad also demonstrated the most stable learning curves over the training.

An important caveat in this comparison of optimizers is that the learning rate of 0.415 was found from a hyperparameter search where Adagrad was used, which could bias the results in favor of Adagrad. To reduce that bias, a lower learning rate of 0.01 was also used for comparison (see Figure \ref{fig:opt}b). With the lower learning rate, both RMSprop and Nadam reported lower training losses than Adagrad, but the losses for all three optimizers remained within an order of magnitude of each other. Furthermore, Adagrad remained competitive when comparing the validation loss. For this reason, Adagrad is a reasonable choice as an optimizer. Improved results might be achieved, however, by including the optimizer type in the hyperparameter search, along with the layer architecture and learning rate.

\bibliographystyle{unsrt}
\bibliography{references}

\end{document}